\documentclass[sn-apa]{sn-jnl}

\usepackage{graphicx}%
\usepackage{multirow}%
\usepackage{amsmath,amssymb,amsfonts}%
\usepackage{amsthm}%
\usepackage{mathrsfs}%
\usepackage[title]{appendix}%
\usepackage{xcolor}%
\usepackage{textcomp}%
\usepackage{manyfoot}%
\usepackage{booktabs}%
\usepackage{algorithm}%
\usepackage{algorithmicx}%
\usepackage{algpseudocodex}%

\usepackage{listings}%

\begin{document}

\title[Article Title]{Optimization of Collective Bayesian Decision-Making in a Swarm of Miniaturized Vibration-Sensing Robots}
%


\author*[1]{\fnm{Thiemen} \sur{Siemensma}}\email{t.j.j.siemensma@rug.nl}




\author*[1]{\fnm{Bahar} \sur{Haghighat}}\email{bahar.haghighat@rug.nl}

\affil*[1]{\orgname{University of Groningen}, \city{Groningen}, \postcode{9747 AG}, \country{The Netherlands}}





\abstract{
Inspection of infrastructure using static sensor nodes has become a well established approach in recent decades. In this work, we present an experimental setup to address a binary inspection task using mobile sensor nodes. The objective is to identify the predominant tile type in a \(1\text{m} \times 1\text{m}\) tiled surface composed of vibrating and non-vibrating tiles. A swarm of miniaturized robots, equipped with onboard IMUs for sensing and IR sensors for collision avoidance, performs the inspection. The decision-making approach leverages a Bayesian algorithm, updating robots' belief using inference. The original algorithm uses one of two information sharing strategies. We introduce a novel information sharing strategy, aiming to accelerate the decision-making. To optimize the algorithm parameters, we develop a simulation framework calibrated to our real-world setup in the high-fidelity Webots robotic simulator. We evaluate the three information sharing strategies through simulations and real-world experiments. Moreover, we test the effectiveness of our optimization by placing swarms with optimized and non-optimized parameters in increasingly complex environments with varied spatial correlation and fill ratios. Results show that our proposed information sharing strategy consistently outperforms previously established information-sharing strategies in decision time. Additionally, optimized parameters yield robust performance across different environments. Conversely, non-optimized parameters perform well in simpler scenarios but show reduced accuracy in complex settings.
}




\maketitle

\section{Introduction}\label{sec:introduction}
Inspection and monitoring are valuable tools for enhancing safety and extending service lifetimes of infrastructure. As a result, automated infrastructure inspection has gained significant interest over recent decades \citep{Lee2023SurveyInspection}. Academic studies highlight applications in areas such as wind turbine, ship hull, agricultural, and infrastructural inspections \citep{Liu2022ReviewTurbines,Schmidt2013ClimbingTechnologies,Carbone2018SwarmAgriculture,Lee2023SurveyInspection}. Industrial examples are the Amazon Monitron \citep{Amazon2024AmazonMonitron} and Fluke 3561 \citep{Fluke2024FlukeGateway}. During automated inspections, sensors replace human perception, detecting damage through indicators such as strain, vibration, temperature, and sound \citep{Gharehbaghi2022APerspectives}. Accordingly, specific sensors are available for a wide range of monitoring and inspection tasks. Vibration sensing is commonly used to detect infrastructure damage by means of extracting features such as natural frequencies, mode shapes, and modal curvature \citep{Farrar2007AnMonitoring, Gharehbaghi2022APerspectives,Doebling1996DamageReview}. 
Damage identification often relies on data driven methods due to the accurate models required for model based approaches \citep{Doebling1996DamageReview}. Within the data driven framework, \citet{Rytter1993VibrationalStructures} outlines a hierarchy of damage detection levels: existence (level 1), location (level 2), severity (level 3), and consequences (level 4). Success implementations at various levels of difficulty are found in both experimental \citep{Alves2023AnSelection, Manson2003ExperimentalWing, Manson2003ExperimentalAircraft, Worden2003ExperimentalStructure, Cui2018AExcitation} and real-world applications \citep{Kaloop2015Stayed-CableMeasurements, Maeck2001DamageMonitoring}. However, challenges persist regarding spatial density, data transfer rates, security, and sensor placement in networks \citep{Mustapha2021SensorReview}. To address the sensor placement problem, \citet{Guo2009Flexure-basedDetection} and \citet{Zhu2009AMonitoring} propose mobile vibration sensors. However, their methods rely on centralized control and the hardware is far from the industry examples mentioned before.

Recent advancements in hardware miniaturization, swarm robotics, and remote monitoring make automated inspections using miniaturized vibration sensing robotic systems a viable solution \citep{Dorigo2021SwarmView}. Swarms of miniaturized robots can serve as mobile sensor carriers, theoretically achieving unlimited spatial resolution, eliminating the problem of sensor placement \citep{Mustapha2021SensorReview}. 
As outlined in \citet{Brambilla2013SwarmPerspective}, swarms are capable of various collective behaviors to solve tasks such as decision-making and localization. These tasks align closely with level 1 and level 2 of damage detection. When designing such artificial swarm behaviors, inspiration is drawn from natural swarms, such as schools of fish or herds of sheep \citep{Schranz2020SwarmApplications}. Within this design approach, no centralized control exists and individuals solely rely on local interactions to produce global behaviors that solve complex tasks. These tasks are studied widely in simulation and experiments, with applications in areas such as space exploration \citep{Nguyen2019Swarmathon:Exploration}, environmental and industrial monitoring \citep{Duarte2016EvolutionRobots,Correll2009MultirobotMachinery}, and agriculture \citep{Albani2017FieldSwarms,Magistri2019UsingSwarms}. However, real-world deployment such as inspection of infrastructure remains limited. 

This paper aims to show the potential of miniaturized swarm robotics for vibration-based infrastructure inspection by leveraging collective decision making.
We do so by studying an binary inspection problem using vibration signals. Specifically, we focus on the \textit{best-of-n} problem as introduced by \citet{Valentini2017ThePerspectives}. In this problem, a collective of robots must determine the best quality option among $\textit{n}$ available options. We focus on a simplified version of this problem, where $n$ equals two. This type of \textit{collective perception} was introduced by \citet{Morlino2010CollectiveRobots}, formalized by \citet{Valentini2016CollectiveSwarm}, and is typically studied in environments of black and white tiles. Without loss of generality, the swarm is tasked to classify the \textit{fill ratio} of white tiles as above or below 0.5. A consensus is reached when all robots in the swarm favor the same final decision. We use vibrating and non-vibrating tiles as a replacement for the white and black tiles.
In short, the swarm is tasked with determining whether the majority of the tiles on the surface is vibrating or non-vibrating. This binary decision does not directly relate to presence of damage, however, signal features such as peak magnitude, root-mean-square (RMS), and variance are used in data driven damage detection methods \citep{Alves2023AnSelection}. 

A variety of collective decision making strategies have been developed to solve the collective perception task. Common strategies are the Direct Modulation of Majority-based Decisions (DMMD) by \citet{Valentini2015EfficientTrade-Off} and Direct Modulation of Voter-based Decisions (DMVD) by \citet{Valentini2014Self-organizedModel}. Moreover,  \citet{Valentini2016CollectiveSwarm} introduce Direct Comparison (DC) as a benchmark algorithm for collective perception where robots compare and adopt the highest-quality opinion from randomly selected neighbouring robots. Methods grounded in Bayesian statistics are studied in \citet{Shan2021DiscreteSharing,Shan2020CollectiveTesting} and \citet{Ebert2020BayesSwarms}. In the latter, robots function as Bayesian modelers, exchanging information with other robots based on two information sharing strategies. The robots either (i) continuously broadcast their ongoing binary observations (\textit{no feedback}) or (ii) continuously broadcast their irreversible decisions once reached (\textit{positive feedback}), pushing the swarm to consensus.

This work is an extension to our work presented in \citet{Siemensma2024CollectiveInspection}, where we study the decision making algorithm proposed by \citet{Ebert2020BayesSwarms} in the aforementioned surface inspection task. We introduced a novel information-sharing strategy called \textit{soft feedback}, to accelerate the decision making. We then developed a calibrated simulation within the robotic simulator Webots, that modeled the real robot behavior. This simulation was used to optimize algorithm parameters via a noise-resistant Particle Swarm Optimization (PSO). Finally, we evaluated the performance of all three information sharing strategies in simulation and through 10 experimental trials.

This extended work builds upon our previous work in several ways. First, we extend the soft feedback strategy and study the effect of various parameter settings. Second, we improve our optimization by considering an extended version of the noise resistant Particle Swarm Optimization (PSO). Third, the calibration between simulated and real robot behaviors has been significantly improved. In our simulation, we include more features as observed in real experiments and quantify the matching. Moreover, we incorporate sampling noise similar to real world sampling noise, reducing the simulation-to-reality gap. Finally, we largely extend our studies in simulations and real-world experiments for the three information sharing strategies: no feedback ($u^-$), positive feedback ($u^+$), and soft feedback ($u^s$). In the simulation, we examine the effects of varying swarm sizes, environmental difficulties, and optimized versus non-optimized parameter sets. To validate the robustness of our findings in simulation, we extend our real experimental studies to various swarm sizes.

This paper is organized as follows: Section \ref{sec:problem_statement} and \ref{sec:related_works} describe the problem and related works. The collective decision making algorithm along with the three information sharing strategies are described in Section \ref{sec:algorithm}. In Section \ref{sec:exp_sim_setup} the robotic platform, simulated environment, and experimental setup are described. The optimization framework is detailed in Section \ref{sec:sim_opt_frameworks}. Finally, the results are presented in Section \ref{sec:exp_results}.
\begin{figure}[t]
    \centering
    \begin{minipage}[b]{0.45\textwidth}
        \centering
        \includegraphics[width=\textwidth]{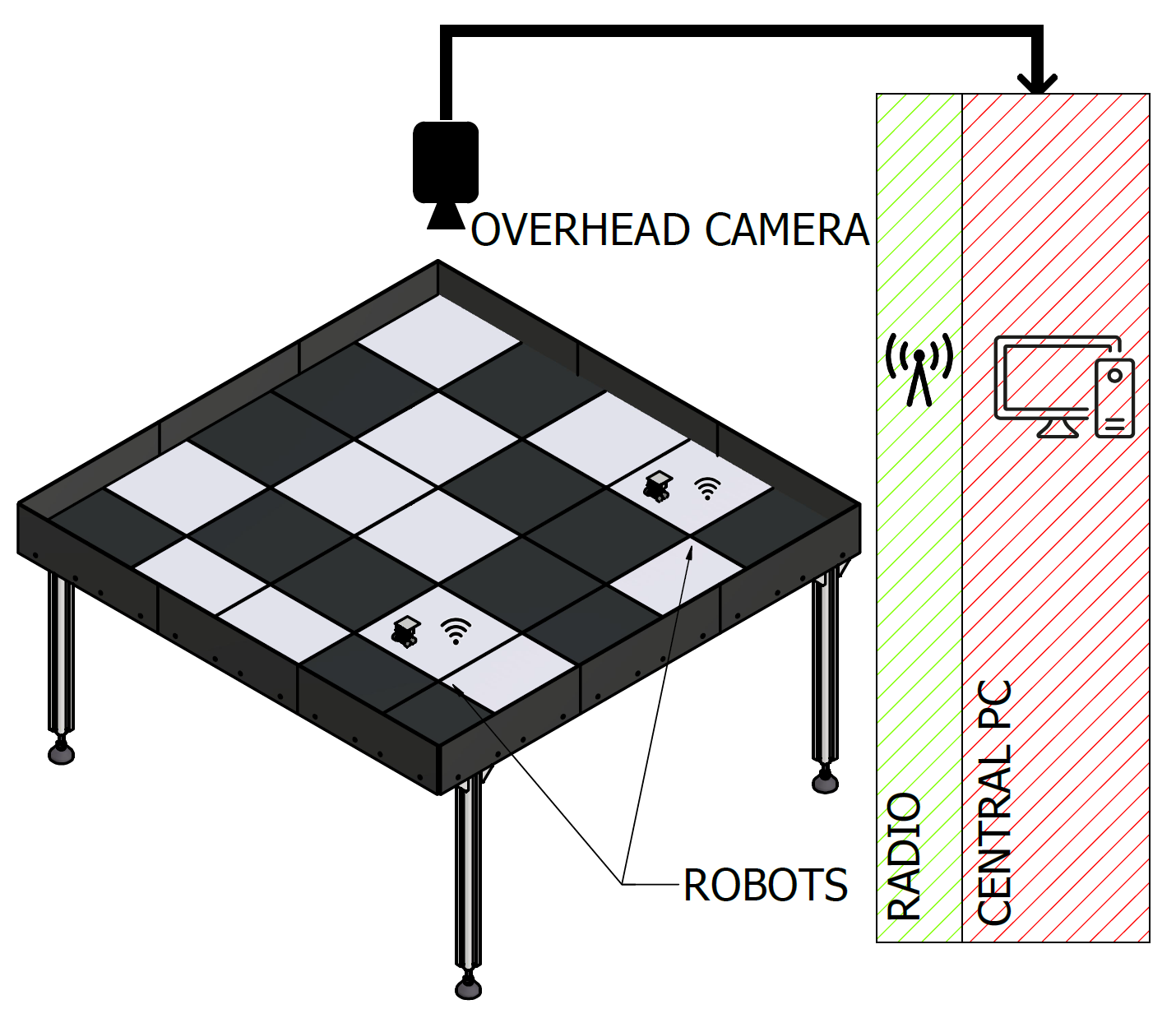}
        (a) Overall experimental setup. 
    \end{minipage}
    \hspace{0.05\textwidth}
    \begin{minipage}[b]{0.375\textwidth}
        \centering
        \includegraphics[width=\textwidth]{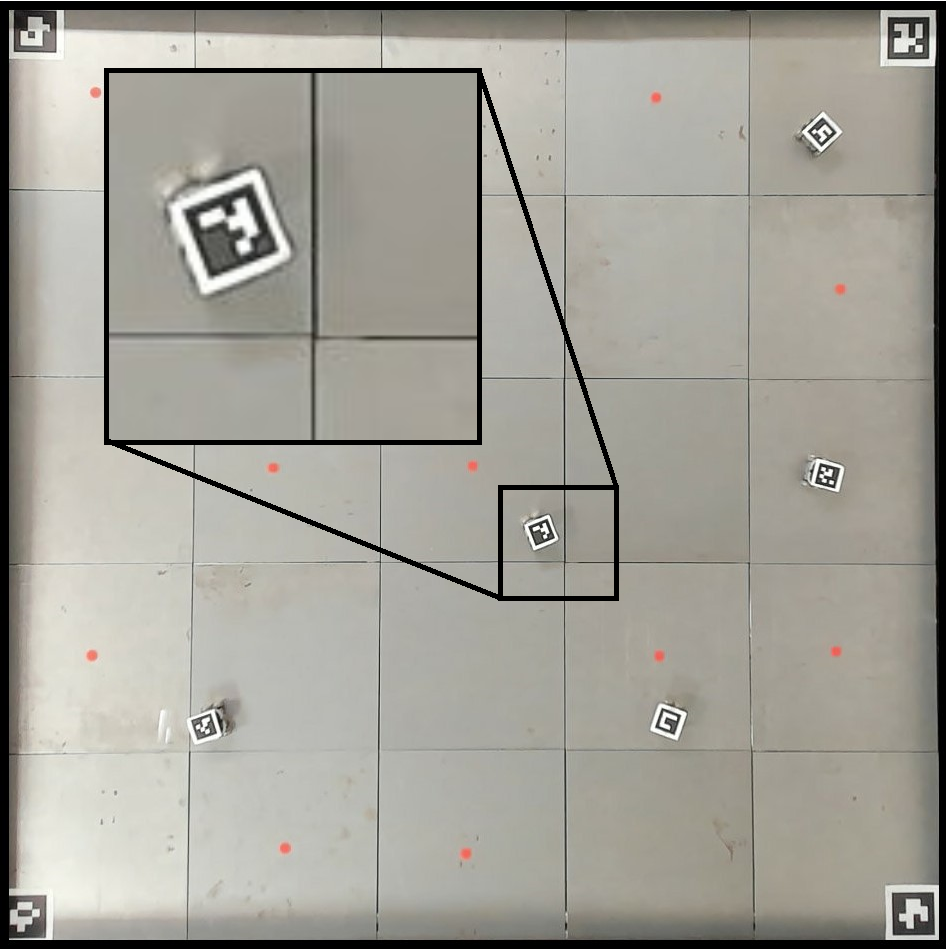}
        (b) View from overhead camera.
    \end{minipage}
\caption{The experimental setup with a fill ratio of $f = \frac{12}{25} = 0.48$. (a) Schematic overview of the setup is shown. The central PC uses the radio and camera data for analysis. Vibration-motors are attached on the bottom side of white tiles. (b) A snapshot from the overhead camera with detailed view (black square). The red dot markings indicate vibrating tiles. Each robot carries a unique AruCo marker for tracking. AruCo markers in the corners of the environment mark the boundaries.}
\label{fig:experimental_setup}
\end{figure}
\section{Problem Statement}\label{sec:problem_statement}
We consider a 2D surface environment, which holds a binary feature. The environment comprises a tiled surface with $N_r \times N_c = N$ tiles and is described by the matrix $E$ as:
\begin{equation}
    \label{eq:environment}
    E := \begin{bmatrix}
    T_{11} & \cdots & \cdots & T_{1N_c} \\
    \vdots & \ddots & \ddots & \vdots \\
    \vdots & \ddots & \ddots & \vdots \\
    T_{N_r1} & \cdots & \cdots & T_{N_rN_c}
    \end{bmatrix},
\end{equation}
where each tile $T_{ij} \in \{0,1\}$, represents the local value of the overall feature. We define the indicator function \( \mathbf{1}_{T_{ij} = 1} \) as:
\begin{equation}
    \label{eq:indicator_function}
        \mathbf{1}_{T_{ij} = 1} = 
    \begin{cases}
    1 & \text{if } T_{ij} = 1 \\
    0 & \text{otherwise}
    \end{cases}.
\end{equation}
The fill-ratio $f \in \{0,1\}$, i.e., the global feature of interest, is given as:
\begin{equation}
    \label{eq:fill_ratio}
    f := \frac{1}{N} \sum_{i=1}^{N_r} \sum_{j=1}^{N_c} \mathbf{1}_{T_{ij} = 1}.
\end{equation}
All robots individually inspect the surface to gather local information. They share this information with the other robots in the swarm to estimate the overall fill-ratio. The goal of each robot is to come up with a final decision $d_f$ that identifies whether the fill-ratio is either above or below $f=0.5$. We denote the correct final decision by $d_f^*$. The inspection task is finished when each robot made a final decision. This problem is hard for fill-ratios close to $f=0.5$ and easier for fill-ratios close to 0 and 1. The difficulty lies in achieving low decision times and high accuracies. Our real experimental setup is a $5 \times 5$ tiled surface as depicted in Figure \ref{fig:experimental_setup}, thus $N$ is limited to 25 in real experiments.

\begin{figure}[t]
    \centering
    \includegraphics[width=0.9\linewidth]{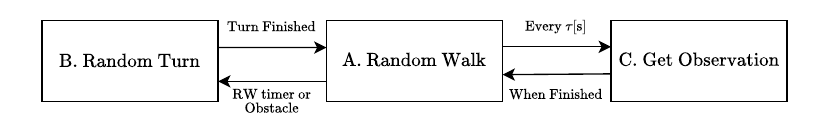}
    \caption{Finite state machine of robot. The robot alternates between the states random walk (exploration), random turn (collision avoidance), and observation (sensing).}
    \label{fig:finite_state_machine}
\end{figure}

 In order to solve this problem, the robots employ a Finite State Machine (FSM) as depicted in Figure \ref{fig:finite_state_machine}. Each robot performs a random walk by driving forward for a random amount of time. When this time expires, or during collision avoidance, the robot turns a random angle. During the random walk the robot acquires samples locally on fixed sampling intervals. This information is shared with all the robots in the swarm in order to collectively estimate the fill ratio.
\section{Related Works}\label{sec:related_works}
Various algorithms have been developed to solve the collective perception problem. The Direct Comparison (DC) algorithm, introduced by \citet{Valentini2016CollectiveSwarm}, is a widely used benchmark for collective decision making. In DC, robots compare the quality of their own opinion with that of randomly selected neighbors and adopt the opinion with the highest quality. Similar methods, such as Direct Modulation of Majority-based Decisions (DMMD) \citep{Valentini2015EfficientTrade-Off} and Direct Modulation of Voter-based Decisions (DMVD) \citep{Valentini2014Self-organizedModel}, do not share opinion qualities directly. Instead, the duration of information sharing is proportional to the opinion quality, a strategy observed in natural swarms as well \citep{Franks2002InformationInsects}.
Some algorithms for collective perception leverage Bayesian statistics. This approach is also used in sensor networks for consensus achievement, inspection, and monitoring \citep{Mustapha2021SensorReview,Makarenko2006DecentralizedNetworks,Alanyali2004DistributedNetworks}. In the context of collective perception, \citet{Shan2020CollectiveTesting} and \citet{Shan2021DiscreteSharing} have applied Bayes’ rule to compute normalized likelihoods for each option (black vs. white), which then represent option quality. 
The algorithm used in this work was introduced by \citet{Ebert2020BayesSwarms}.
The authors propose Bayesian inference on a Beta-distribution that represents the collective estimate on the fill-ratio. The robots integrate personal observations with messages from other robots into their posterior distribution. By evaluating the Cumulative Density Function (CDF) of the Beta-distribution the robots compute their current opinion on the fill ratio (belief). Two methods of feedback between robots are considered: (i) no feedback, i.e. always broadcasting current observations and (ii) positive feedback, broadcasting final decisions when made. The authors found that exploiting positive feedback accurately accelerated decision making. Moreover, increasing the time between observations increases the decision accuracy. This result is confirmed by other studies \citep{Bartashevich2019PositiveDecision-making,Shan2020CollectiveTesting,Bartashevich2021Multi-featuredCorrelations}. It shows that spatial correlation of samples is detrimental for accurate decision making. \citet{Bartashevich2019BenchmarkingDecision-making} propose difficulty measures for collective perception. They suggest that the difficulty not only lies in fill-ratio, but rather the spatial correlation and clustering of tiles within the environment determine the difficulty of collecting representative samples. 
The spatial correlation in the environment can be defined by the Moran Index as:
\begin{equation}
    \label{eq:moran_index}
    E_{MI} = \frac{N}{\sum_{ij}^N\omega_{ij} } \cdot \frac{\sum_{ij}^N\omega_{ij}(c_{i} - f)(c_{j} -f )}{ \sum_{i}^N( c_i - f)^2},
\end{equation}
where the value of $\omega_{ij} = \omega_{ji}$ equals $1$ if tile $i$ and $j$ have a common side (top, left, right, and bottom), and $0$ otherwise. Moreover, $f$ is the overall fill-ratio, and $c_i,c_j$ equal the values of the flattened matrix $E$ at location $i$ and $j$ respectively. The Moran Index approaches $1$ for high correlations, and $-1$ for no correlation, whereas randomized floor patterns have a value close to $0$. Second, to measure the clustering of tiles, the entropy of the environment $E$ is computed as:
\begin{subequations}
    \label{eq:entropy}
    \begin{equation}
        E_e = \frac{H_c^{max} - H_c}{H_c^{max}}
    \end{equation}
    \begin{equation}
        H_c = - \sum_{i=1}^M \frac{|C_i|}{N_{v}} \log_2\frac{|C_i|}{N_{v}}
    \end{equation}
        \begin{equation}
        H_c^{max} = - \sum_{N_{v}} \frac{1}{N_{v}} \log_2\frac{1}{N_{v}},
    \end{equation}
\end{subequations}
where $M$ is the number of clusters $C_i$ in $E$, $|C_i|$ the number of white tiles in cluster $C_i$, and $N_{v}$ the total number of white tiles ($N_{v} = N\cdot f$). $E_e$ approaches $0$ for a checkerboard-like pattern and $1$ for a monolithic pattern.
In conclusion, the challenges in collective perception stem from the environmental features in terms of fill ratio, spatial correlation, and clustering, as well as the search behavior of the robots, which can be optimized. In our previous work, we took first steps in optimizing such robot behaviors, largely based on the simulation frameworks developed by \citet{Chiu2024OptimizationOptimization} and \citet{Haghighat2022AnRobots}.
\section{Algorithm} \label{sec:algorithm}
We adopt the algorithm proposed by \cite{Ebert2020BayesSwarms} and investigate various information sharing strategies. The overall structure is given in Algorithm \ref{algo:bayesianalgorithm}. Each robot inspects the environment locally by gathering observations at fixed observation intervals. 
The exploration behavior of the swarm is characterized by a Levy-flight type random walk on the 2D surface. Each robot moves forward for $T_r$ milliseconds of time, sampled from a Cauchy distribution as:
\begin{equation}
    T_r \sim \text{Cauchy}(\gamma_0, \gamma),
\end{equation}
where \( \gamma_0 \) is the location parameter (mode) and \( \gamma \) is the scale parameter, representing the average absolute deviation from the mode.
When the time $T_r$ expires, the robot turns a random angle $\phi$ as:
\begin{equation}
    \label{eq:random_angle}
    \phi \sim U(-\pi,\pi),
\end{equation}
in the direction of sgn($\phi$). When the robot detects an obstacle within a $\theta_c$[mm] range, it enters collision avoidance mode. In this mode, the robot randomly adjusts its orientation by turning a random angle, as described in Equation \eqref{eq:random_angle}. 
\begin{algorithm}[t]
\caption{Collective Bayesian Decision Making}\label{algo:bayesianalgorithm}
\begin{algorithmic}[1]
\State \textbf{Inputs: }$T_{end},O_c,\theta_c,\tau,p_c$
\State  \textbf{Init: }$\alpha = 1, \beta =1,d_f =-1, \bar O_c = 0,t_s =0,T_r$
\While{$t < T_{end}$}
\State Perform random walk for $T_r$ time
\If{$t - t_s > \tau$}
    \State $O\gets $ \textit{getObservation()}\Comment{Get binary observation}
    \State $t_s\gets t$ \Comment{Observation timestamp}
    \State $\text{Beta}(\alpha,\beta) \gets \textrm{Beta}(\alpha + O, \beta+ (1- O)) $ \Comment{Update modeling of $f$}
    \State $p \gets P(\text{Beta}(\alpha,\beta)<0.5) $\Comment{Update belief of $f$}
    \State $\bar O_{c} \gets \bar O_{c}+1 $ \Comment{Observation count}
\EndIf
\If{$(u^s \text{ \textbf{or} } d_f == -1)$  \textbf{and} $ (\bar O_{c} > O_c)$}
    \If{$p>p_c$}
        \State $d_f \gets 0$
    \ElsIf{$(1-p) > p_c $}
     \State $d_f \gets 1$
    \EndIf
\EndIf
\State $m\leftarrow \text{\textit{constructMessage($p,O,d_f$)}}$\Comment{Prepare message}
\State \textit{Broadcast($m$)}\Comment{Broadcast message}
\If{Messages in queue}
    \For{{Message in Messages}}
        \State $m \gets $ Message \Comment{Receive message from swarm}
        \State $\text{Beta}(\alpha,\beta) \gets \textrm{Beta}(\alpha + m, \beta+ (1- m))$ 
        \Comment{Update modeling of $f$}
        \State $\bar O_{c} \gets \bar O_{c}+1 $ \Comment{Observation count}
      \EndFor
\EndIf
\EndWhile
\end{algorithmic}
\end{algorithm}
During exploration, every $\tau$ seconds, the robot acquires a local binary sample on the surface as being \textit{vibrating} or \textit{non-vibrating}. In our previous work \citep{Siemensma2024CollectiveInspection}, the timer \( \tau \) was inactive during collision avoidance. In this extended work, we keep the timer \( \tau \) active at all times to simplify the algorithm, though this increases spatial correlation and, consequently, the difficulty of the task. Assuming sampling without noise, the probability of observing vibrating and non-vibrating are denoted by $f$ and $(1-f)$ respectively. Each sample can be seen as drawn from a Bernoulli distribution with probability $f$, i.e, the proportion of \textit{vibrating} tiles in the overall setup as:
\begin{equation}
    O \sim \textrm{Bernoulli}(f).
\end{equation}
The fill-ratio $f$ is unknown by the robots in the swarm and modeled by a Beta-distribution with a prior defined as: 
\begin{equation}
    f \sim \textrm{Beta}(\alpha,\beta),
\end{equation}
where $\alpha,\beta$ denote vibrating and non-vibrating samples respectively. We obtain the posterior distribution by integrating personal observations as:
\begin{equation}
    f \ | \ O \sim \textrm{Beta}(\alpha + O, \beta+ (1- O)).
\end{equation}
Similarly, we integrate messages $m$ from other robots into the posterior as: 
\begin{equation}
    f \ | \ m \sim \textrm{Beta}(\alpha + m, \beta+ (1-m)).
\end{equation}
Each robot holds a belief $p$ about the state of the surface, i.e. being mostly vibrating or non-vibrating, by evaluating the Beta distribution's CDF at 0.5 as:
\begin{equation}
    p := P(f<0.5),
\end{equation}
where $p$ denotes the probability that the surface is mostly non-vibrating. Each robot makes a final decision by evaluating $p$ against a pre-defined threshold $p_c$ as:
\begin{equation}
    d_f \leftarrow \begin{cases}
        0 \quad \text{if } p \geq p_c\\
        1 \quad \text{if } p \leq (1-p_c),\\
    \end{cases}
\end{equation}
where $d_f$ denotes the final decision of the robot. When $d_f=0$ the final decision corresponds to the surface being mostly non-vibrating. Conversely, $d_f=1$ corresponds to the surface being mostly vibrating. To avoid premature decisions, each robot is only allowed to make a final decision after $O_c$ posterior updates. 

In real world experiments, each observation $O$ is subject to noise. We observe each tile in the real experimental setup via a simple strategy as described in Algorithm \ref{algo:inspection_logic}. Before each observation, the robot will pause for 500 ms to stabilize. Afterwards, the robot acquires acceleration values at a sampling rate of 350 Hz for $T_s$ seconds. In this work, we set $T_s$ equal to 500 ms, bringing the total pause time for every observation to 1000 ms. 
The DC component of each signal is removed by employing a first-order high-pass filter with cutoff frequency $\omega_n = 40$ Hz. Given a sampling rate of $350$ Hz, the filter parameters $\alpha_1$, $\alpha_2$, and $\alpha_3$ in Equation \eqref{eq:high_pass_filter} are configured to values: $0.20$, $0.60$, and $-0.60$ respectively. We define the filtered signal at time step $i$ as $\hat a_i$:
\begin{equation}
    \label{eq:high_pass_filter}
    \hat a_i :=  \alpha_1 \hat a_{i-1} +\alpha_2 a_i  +\alpha_3 a_{i-1},
\end{equation}
where $a_i$ is the magnitude of the IMU's raw acceleration data $a_x$, $a_y$, and $a_z$. The Root-Mean-Square (RMS) of $\hat a$ returns the energy of the signal as $\hat E =\sqrt{\frac{1}{n}\sum_{i=1}^n \hat a_i^2}$. Subsequently, the observation $O$ is determined by comparing $\hat E$ with a threshold $\theta_E$ that distinguishes vibrating from non-vibrating tiles as:
\begin{equation}
\label{eq:observation}
    O=\begin{cases}
            1& \text{if } \hat E >\theta_E\\
             0& \text{if } \hat E \leq \theta_E.
        \end{cases}\\
\end{equation}
\begin{algorithm}[t]
    \caption{Surface Inspection Logic: \textit{getObservation()}}\label{algo:inspection_logic}
\begin{algorithmic}[1]
\State \textbf{Init: }$n=0,\theta_E=1.55,T_s = 500$[ms]
\State Pause(500ms)
\While{$t<T_s$}
    \State $a_x,a_y,a_z \leftarrow $ \textit{get acceleration values}
    \State $a_n =  \sqrt{a_x^2 +a_y^2 +a_z^2 }$
    \State $\hat a_n:=\alpha_1 \hat a_{n-1} + \alpha_2 a_n + \alpha_3 a_{n-1} $\Comment{High-pass filter}
    \State $n \leftarrow n +1$
\EndWhile
\State $\hat E  = \sqrt{\frac{1}{n} \sum_{i=0}^n \hat a_i^2}$\Comment{RMS}
\If{$\hat E > \theta_E$}
\State $O\leftarrow 1$
\ElsIf{$\hat E \leq \theta_E$}
\State $O \leftarrow 0$
\EndIf
\end{algorithmic}
\end{algorithm}
\subsection{Feedback Strategies}\label{subsec:info_sharing}
In addition to environmental difficulty, decision time and accuracy are strongly influenced by the swarms' information sharing strategy. Each robot broadcasts a message subsequent to each observation. In the work by \citet{Ebert2020BayesSwarms}, only nearby robots are considered. In this work, we use an all-to-all communication strategy. The swarm employs one of three feedback strategies, as outlined in Algorithm \ref{algo:info_sharing_strategy}. In \citet{Ebert2020BayesSwarms}, the no feedback ($u^-$) and positive feedback ($u^+$) strategies are proposed. We propose a new strategy, termed soft feedback ($u^s$) to accelerate the decision making.
\begin{algorithm}[t]
\caption{Feedback Strategies: \textit{constructMessage()}}\label{algo:info_sharing_strategy}
\begin{algorithmic}[1]
\State \textbf{Inputs: }$p$, $O$, $d_f$
\State \textbf{Output: } $m$
\If{$u^s$}
 \State $\Gamma \gets \text{Var(Beta)}$\Comment{Variance on Beta distribution}
 \State $\delta = e^{-\eta \Gamma} |\frac{1}{2} - p|^\kappa$
 \State $m \leftarrow \text{Bernoulli}(\delta \cdot (1-p) + (1-\delta) \cdot O)$
\ElsIf{($u^+$) \textbf{and} ($d_f \neq -1$)}
\State $m\leftarrow d_f$
\Else
\State $m\leftarrow O$
\EndIf
\end{algorithmic}
\end{algorithm}
(i) Using no feedback ($u^-$), each robot broadcasts its latest observation $O$. (ii) Using positive feedback ($u^+$), each robot broadcasts its latest observation $O$ until reaching a final decision. From this point in time, the final $d_f$ is shared. The intuition behind the positive feedback strategy is to push the swarm to consensus once individual robots reach a final decision. However, this is not effective when no robot is able to reach a final decision. Moreover, the inspection task is finished upon reaching a final decision, thus positive feedback only has an effect on swarms that have dissimilar belief trends within the swarm. In this case, a decisive robot will push the indecisive robots to the same decision. For an all-to-all communication strategy, beliefs tend to have similar trends, as all robots hold similar Beta distributions. Using such communication, a more effective approach to achieving consensus is to progressively guide the swarm towards the preferred final decision, i.e. the most probable option. To accomplish this, we introduce soft feedback ($u^s$). In this method each robot constructs a message $m$, representing its preferred final decision. This message is randomly sampled with a probability comprised of two elements: the robots long term belief ($p$) and the robots current observation ($O$). The combination of $O$ and $p$ in $m$ is determined by $\delta$, a Bernoulli random variable with a probability distribution defined as:
\begin{equation}
    m \sim \text{Bernoulli}(\delta \cdot (1-p) + (1-\delta) \cdot O),
\end{equation}
where $\delta \in [ 0,1] $. This strategy gradually pushes the swarm to consensus by increasing the influence of $p$ in $m$ as $\delta$ increases. To this end, we incorporate the variance on the Beta distribution in $\delta$, where the variance is defined as:
\begin{equation}
    \Gamma = \frac{\alpha \beta}{(\alpha + \beta)^2 (\alpha + \beta +1)}.
\end{equation}
As a robot gathers increasingly more samples, the variance on its Beta distribution will decrease, thus the estimate gets more accurate. We construct $\delta$ such that it increases with decreasing variance as:
\begin{equation}
    \delta = e^{-\eta \Gamma} |\frac{1}{2} - p|^\kappa ,
\end{equation}
where $e^{-\eta \Gamma} \in (0,1]$ for $\gamma \in \mathbb{R}^+$ and $\eta \in \mathbb{R}^+$. As $\Gamma$ decreases, $e^{-\eta \Gamma}$ will approach 1. The speed of approach is tunable by the parameter $\eta$. 
The term \( |0.5 - p|^\kappa \) in \(\delta\) prevents an increase of $p$ in $\delta$ if $p$ is uncertain. By considering the absolute distance of \(p\) from the indecisive state (i.e., \(p = 0.5\)), \(\delta\) remains close to zero when a robot's belief is indecisive, i.e. \(p \approx 0.5\). This effect can be amplified by increasing the value of \(\kappa\). Moreover, this approach ensures a bound on $\delta$ (i.e., \(\delta \in [0,0.5^\kappa]\)), providing a maximum for the amount of $p$ in each message $m$. In our previous work, \(\kappa\) was fixed at 2 \citep{Siemensma2024CollectiveInspection}. In this work, we optimize the value of \(\kappa\) for our environment.

\section{Experimental and Simulation Setup}\label{sec:exp_sim_setup}
Our experimental setup, shown in Figure \ref{fig:experimental_setup}, is built around (i) a swarm of $3 \text{cm}$-sized vibration-sensing robots and (ii) a $1 \text{m} \times 1 \text{m}$ tiled surface of 25 metallic tiles. This complete setup was first demonstrated in our earlier work \citep{Siemensma2024CollectiveInspection}, here we repeat the main aspects for completeness.
\subsection{Robot}\label{subsec:robot}
\begin{figure}[t!]
    \centering
    \begin{minipage}[t]{0.2\textwidth}
        \centering
        \includegraphics[width=\textwidth]{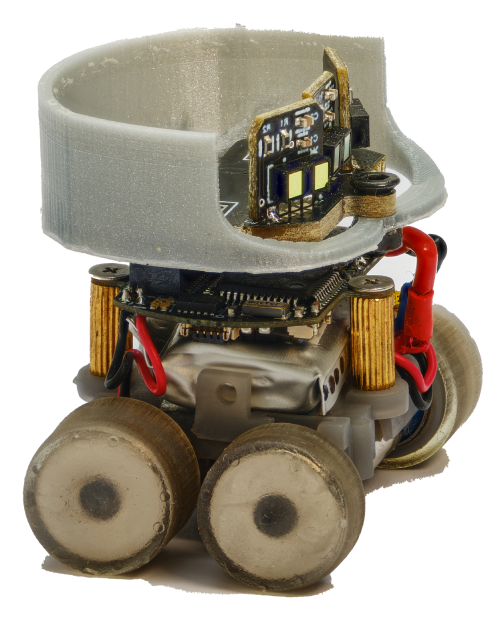}
        (a) Real Robot. 
    \end{minipage}
    \begin{minipage}[t]{0.4\textwidth}
        \centering
        \includegraphics[width=\textwidth]{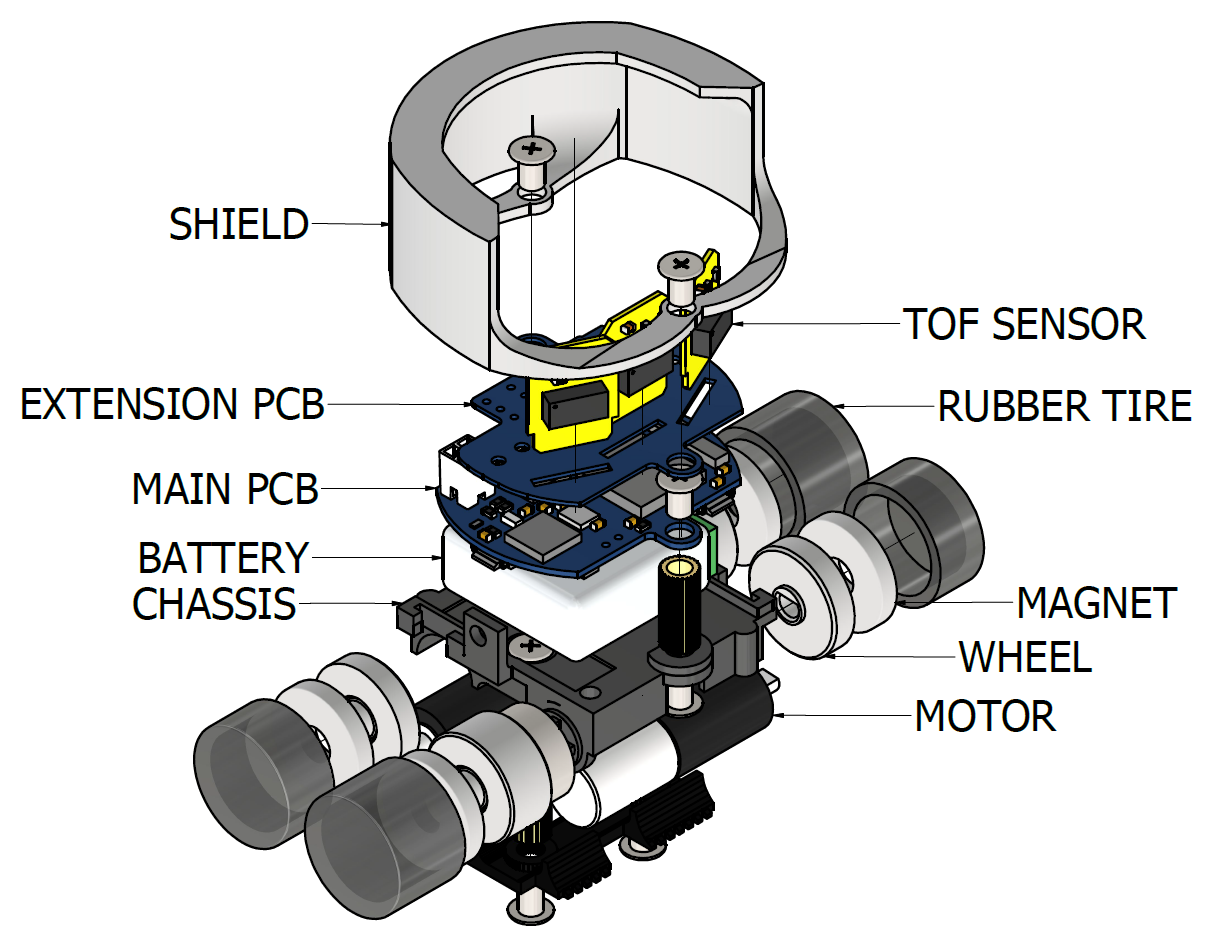}
        (b) Robot CAD view.
    \end{minipage}
    \begin{minipage}[t]{0.225\textwidth}
        \centering
        \includegraphics[width=\textwidth]{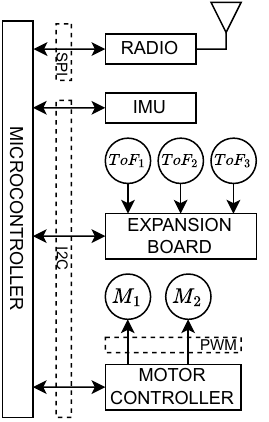}
        (c) Schematics.
    \end{minipage}
\caption{We use a revised and extended version of the Rovable robot.
(a) The extended robot with IR sensor board. (b) Exploded 3D CAD view of the
extended robot. (c) Electronic block diagram of the extended robot. The microcontroller (Atmel SAMD21G18) interfaces with the IMU (MPU6050), 2.4 GHz radio (nRF24L01+), motor-controllers (DRV8835), and ToF IR sensors (VL53L1X).}
\label{fig:vibration_sensing_robot}
\end{figure}

We use an updated version of the Rovable robot introduced by \citet{Dementyev2016Rovables:Wearables} and enhanced in our earlier work \citep{Siemensma2024CollectiveInspection}. As shown in Figure \ref{fig:vibration_sensing_robot}, the robot measures $25\,\text{mm} \times 33\,\text{mm} \times 35\,\text{mm}$ and is equipped with two custom Printed Circuit Boards (PCBs). The first PCB integrates the microcontroller, an Inertial Measurement Unit (IMU), motor controllers, power circuitry, and a radio module. The robot uses the IMU (MPU6050) for angular positioning during turning and for sensing the vibrations on the surface. The radio (RF24) transmits messages up to 144 bytes in a network layer protocol, used for information sharing between robots. Moreover, we use the radio to extract data from the decision making process for post processing. The second PCB is dedicated to collision avoidance and is fitted with infrared (IR) sensors of type VL53L1X. The design of the main PCB remains largely similar to that of the original in \cite{Dementyev2016Rovables:Wearables}, with updates made for revised components. In our previous work 
\citep{Siemensma2024CollectiveInspection}, we developed a second PCB including three Time-of-Flight (ToF) IR sensor modules. These sensors are positioned to cover angles of $0^\circ$, $25^\circ$, and $-25^\circ$ relative to the robot's forward-facing direction, each with a $27^\circ$ field of view and an effective range of up to $1\,\text{m}$. This setup allows for object detection within a $2\times (25+\frac{27}{2}) = 77^\circ$ field of view up to one meter in range. A 3D-printed shield encloses the extension PCB at the rear and sides, improving the robot’s visibility to the IR sensors of neighbouring robots. The average battery life of the robot is at maximum $20$ minutes when continuously driving at full speed. The robot moves on four magnetic wheels, with only one front and rear wheel motorized. The other front and rear wheels are not driven. The motors, controlled by pulse-width modulation (PWM), enable the robot to reach speeds of approximately $5\,\text{cm}$ per second at full power across metallic surfaces. A rubber tire mounted around the magnetic wheels reduces the attractive forces between the magnets of different robots, while maintaining sufficient force to adhere to the surface.

\subsection{Real Environment}\label{subsec:real_environment}
The robots operate in a $1\text{m}\times 1\text{m}$ tiled 2D surface environment demonstrated in our earlier work \citep{Siemensma2024CollectiveInspection}. The surface consist of tiles, each measuring 20 cm by 20 cm, arranged in a 5-by-5 grid. The thickness of each tile is 0.5 mm and the material is S235JR, which is general construction steel that can be expected in structures. Two types of tiles are present: vibrating and non-vibrating. The vibrating tiles are actuated by two miniature vibration motors mounted on top of each other underneath the center of each tile (ERM 3V Seeed Technology motors). For accurate representation of the environment, the sampling strategy requires calibration of the vibration threshold $\theta_E$. Moreover, there is some propagation of vibration between adjacent tiles. This causes noise in our experiments. Each tile is secured to an aluminum frame using 2 cm by 1 cm magnetic tape at the corners, with the frame comprising four central strut profiles and four others along the perimeter. The environment is enclosed by walls of black Plexiglass for the robots to detect the bounds.  For visual tracking, we utilize an overhead camera (Logitech BRIO 4k) and AruCo markers for identification and monitoring of the robots' positions. The camera feed information is not used for controlling the robots.
\begin{figure}[t!]
    \centering
    \begin{minipage}[t]{0.19\textwidth}
        \centering
        \includegraphics[width=0.9\textwidth]{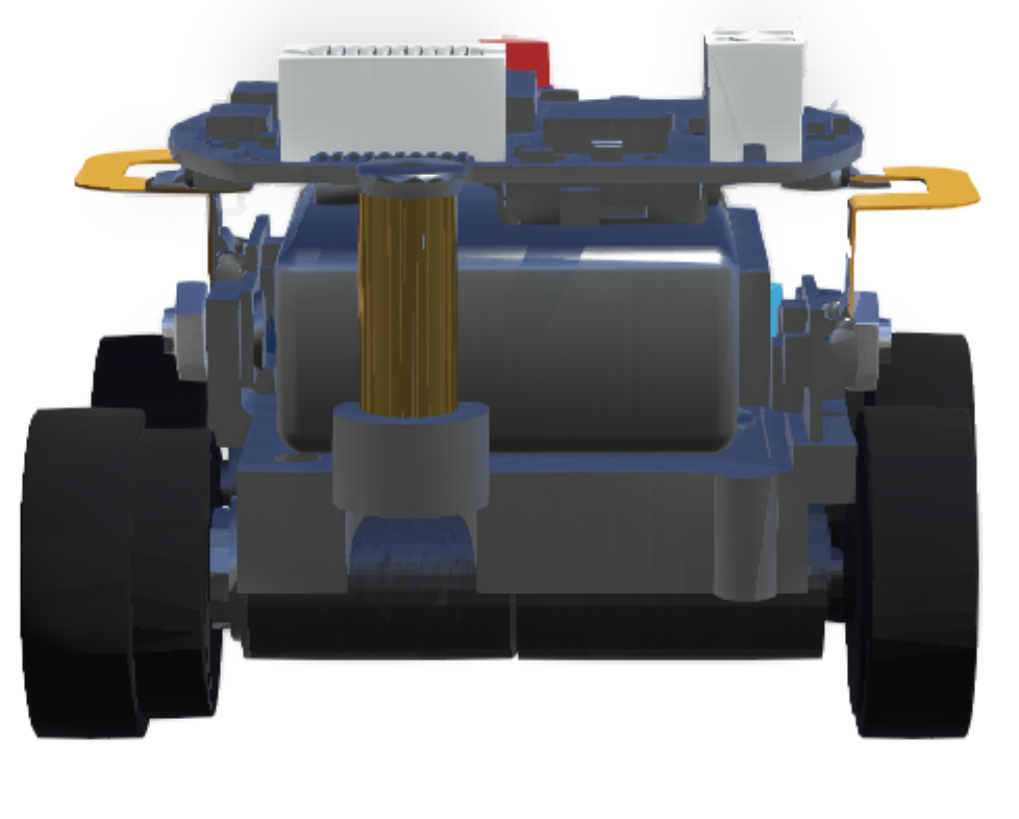}
        (a) Model, front
    \end{minipage}
    \begin{minipage}[t]{0.19\textwidth}
        \centering
        \includegraphics[width=0.9\textwidth]{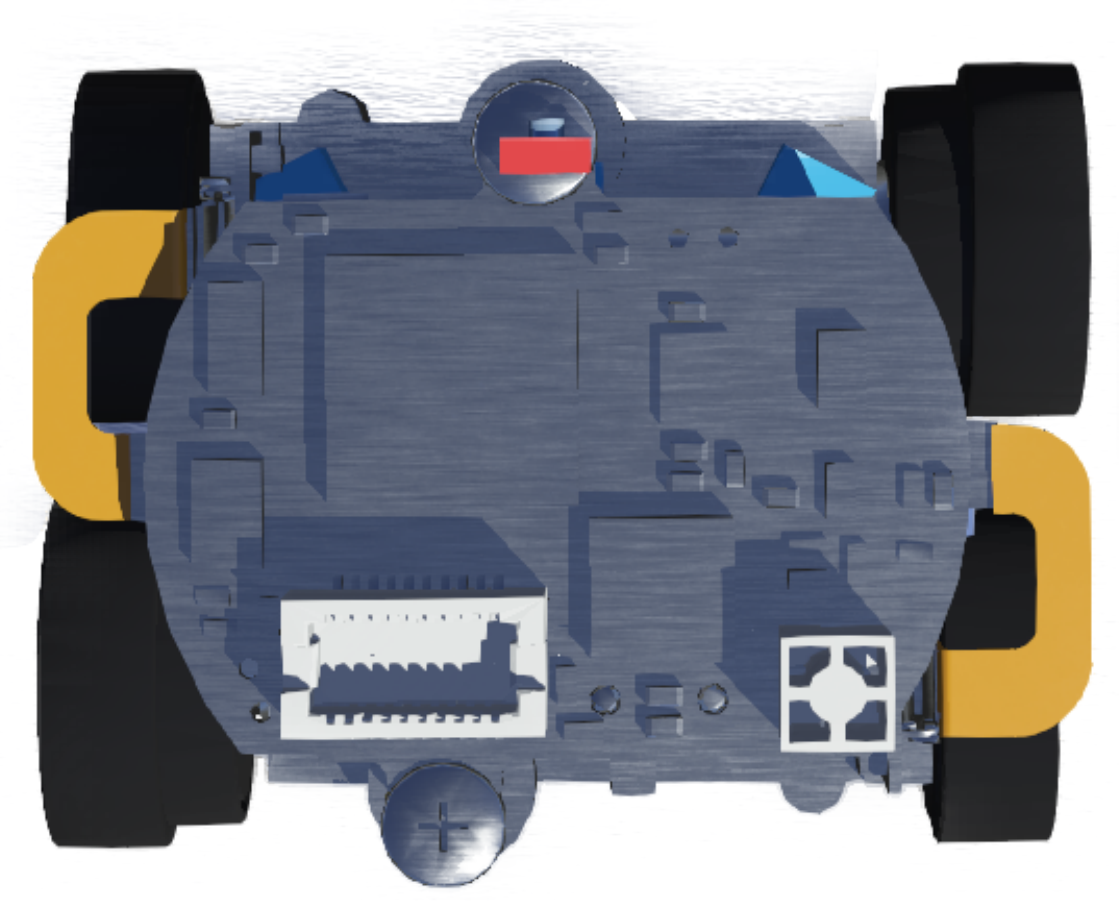}
        (b) Model, top
    \end{minipage}
    \begin{minipage}[t]{0.21\textwidth}
        \centering
        \includegraphics[width=\textwidth]{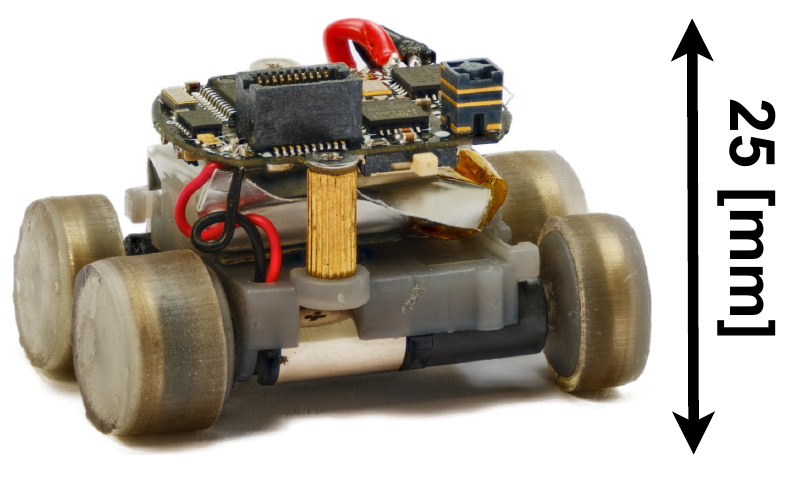}
        (c) Real, front
    \end{minipage}
    \begin{minipage}[t]{0.21\textwidth}
        \centering
        \includegraphics[width=\textwidth]{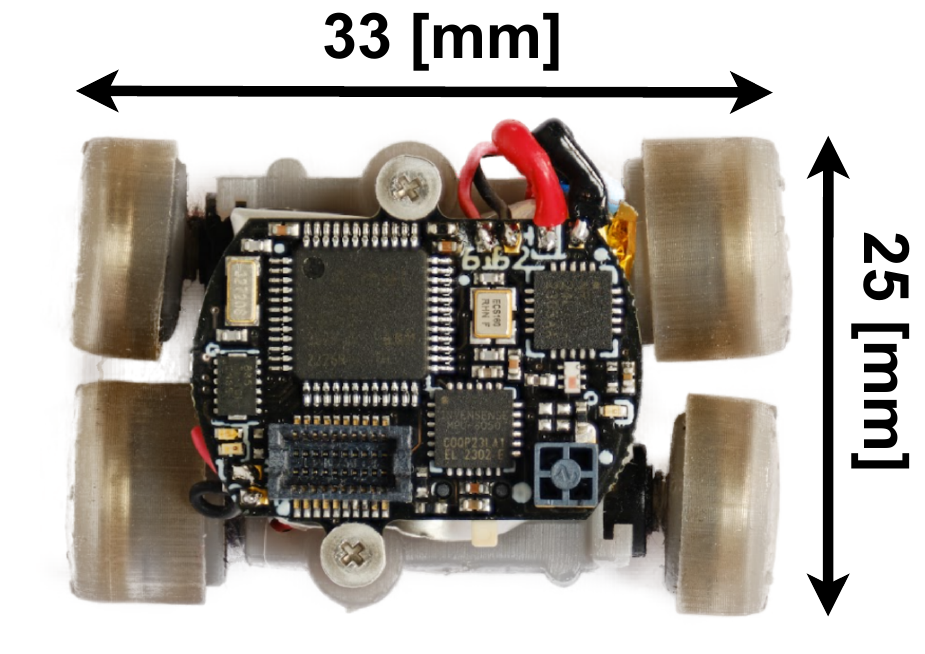}
        (d) Real, top
    \end{minipage}
\caption{We use a simpler robot model in simulation, with IR sensors directly simulated on the main PCB. Our simulated collision avoidance closely matches reality.}
\label{fig:robot_views}
\end{figure}
\subsection{Simulated Environment}\label{subsec:simulated_environment}
Our simulated environment builds upon our previous work 
\citep{Siemensma2024CollectiveInspection} to accurately model the performance of the robots as observed in the real experimental setup. The simulated robot captures the embodiment of the real robot in large detail, as depicted in Figure \ref{fig:robot_views}. In contrast to agent-based simulations, physics-based simulations allow for accurate modeling of robot dynamics during movements and collision avoidance behaviors. Within the simulation, we replicate the state machine as implemented on the real robot, described by Algorithm \ref{algo:bayesianalgorithm}. Each robot performs a random walk, gathers observations at fixed intervals, and performs collision avoidance upon detecting an obstacle within $\theta_c$ [mm]. Due to physical differences among the real robots, we observe variability in the behaviors of robots even when using the same algorithm parameters. This noise is replicated within the simulation to minimize the sim-real gap. 
In our simulation, we incorporate three phenomena to ensure calibration with real-world experiments: (i) Mechanical misalignment, arising from variations in motor performance and chassis mounting; (ii) Battery drainage, reflected in the gradual decline of battery potential and, consequently, driving power over time; and (iii) Observation errors, encompassing false positives (FP) and false negatives (FN) caused by noise in the experimental setup. Replication of these three phenomena is done by stochastically implementing an imperfection within the simulation and manually calibrating it to match observations from the real experimental setup. 

Every robot tends to veer off a straight path due to mechanical misalignment and fabrication imperfection. Moreover, every robot exhibits a different driving speed. To account for these effects, in simulation, we introduce random offsets to the motor speeds, with each robot receiving unique offsets. The left and right motor speeds, denoted as signals \( v_l \in [0, 100] \) and \( v_r \in [0, 100] \), are initialized at a common set speed \( v_s \). For each robot, we then randomly set the motor input speeds as follows:
\begin{subequations}
    \begin{equation}
        v_l = v_s \cdot (1 - m_d) \cdot s_d \cdot b_d
    \end{equation}
    \begin{equation}
        v_r = v_s \cdot (1 + m_d) \cdot s_d \cdot b_d,
    \end{equation}
\end{subequations}
where the terms $(1 \pm m_d)$ account for mechanical misalignment, $s_d$ introduces speed differences among robots, and $b_d$ represents battery voltage drop which decays linearly. $m_d$ is drawn from a uniform distribution $U(a, b)$, while $s_d$ is drawn from a gamma distribution $\text{Gamma}(k, \theta)$. These distributions are chosen empirically based on data collected from the real robots. Furthermore, the battery voltage drop is assumed to be consistent across all robots. 

In our experimental setup, both vibrating samples \((O=1)\) and non-vibrating samples \((O=0)\) are subject to FN and FP observations, respectively. To quantify this, we collect vibration data randomly across our experimental setup and classify them as vibrating or non-vibrating based on camera-captured position data sampled at 5 Hz. Our observations indicate that the distributions of \(O=0\) and \(O=1\) can be accurately modeled using Gamma distributions. Therefore, in simulation, we draw each sampled RMS value $\hat E$ from one of two Gamma distributions, depending on whether the robot observes a vibrating or non-vibrating surface section as:
\begin{equation}
\hat E \sim \begin{cases}
    \text{Gamma}(k_v, \theta_v) \quad &\text{Vibrating tile} \\
    \text{Gamma}(k_{nv}, \theta_{nv}) \quad &\text{Non-vibrating tile}.
\end{cases}
\end{equation}

\section{Optimization Framework}\label{sec:sim_opt_frameworks}
\begin{figure}
    \centering
    \includegraphics[width=0.75\linewidth]{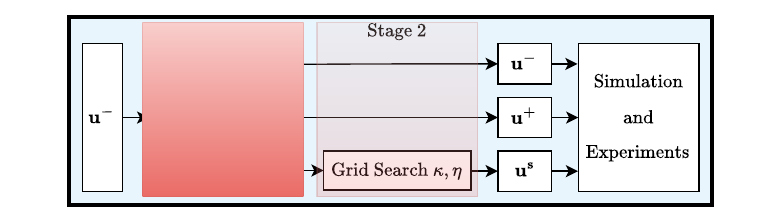}
    \caption{Our optimization and evaluation process: Stage 1 optimizes joint algorithm parameters across three feedback strategies using noise-resistant Particle Swarm Optimization (PSO), utilizing the no-feedback strategy ($u^-$). Stage 2 employs a grid search to optimize soft feedback parameters $\kappa$ and $\eta$. Afterwards, all three feedback strategies are assessed in simulation and experiments.}
    \label{fig:optimization_framework}
\end{figure}
We employ the two-stage optimization approach as depicted in Figure \ref{fig:optimization_framework} to identify the optimal algorithm parameters. In the first stage, we optimize the algorithm parameters ($\gamma_0,\gamma,\tau,\theta_c,O_c$) that are common across the three feedback strategies by employing the no feedback strategy ($u^-$). Given that $u^-$ has no reinforcing feedback, we hypothesize that this optimization will give rise to optimal sampling behavior. To achieve this, we utilize a noise-resistant Particle Swarm Optimization (PSO) method.
In the second stage, we conduct a grid search to optimize the soft feedback parameters, providing valuable insights into the effects of soft feedback with respect to its parameters, $\kappa$ and $\eta$.
To facilitate both optimization processes, we develop a simulation framework within the Webots environment. Each simulation is initiated via a Python script that takes as input the environment and parameters, and returns a fitness score and a detailed log file for analysis.
\subsection{Particle Swarm Optimization}
We use a noise-resistant variant of the PSO method. The velocity and position of particle $i$ are updated at iteration $k$ as:
\begin{subequations}
\label{eq:pso_equations}
\begin{equation}
    \label{eq:pso_vel}
     \mathbf{v}_{i}^{k+1} = \omega \cdot \mathbf{v}_{i}^k + \omega_p \cdot \mathbf{R_1}(\mathbf{p}_{b_{i}} - \mathbf{p}_{i}^k) + \omega_g \cdot \mathbf{R_2}(\mathbf{g}_{b_{}} - \mathbf{p}_{i}^k)
\end{equation} 
\begin{equation}
    \label{eq:pso_pos}
    \mathbf{p}_{i}^{k+1} = \mathbf{p}_{i}^k + \mathbf{v}_{i}^{k+1},
\end{equation}
\end{subequations}
where $\mathbf{v}_{i}^k$ and $\mathbf{p}_{i}^k$ are the velocity and position vector of particle $i$ at iteration $k$. $\mathbf{p}_{b_{i}}$ and $\mathbf{g}_{b_{}}$ represent the position vector of the personal best and global best evaluations for particle $i$, respectively. We set the PSO weights for personal best and global best as $[ \omega_p \ \omega_g ] = [ 0.75 \ 0.75 ]$, balancing local and global exploration \citep{Poli2007ParticleOverview,Gad2022ParticleReview,Innocente2010CoefficientsGuidelines}. We define the inertia weight $\omega$ as decreasing linearly throughout the iterations, starting at $1.0$ and reducing to $0.4$ to enhance convergence \citep{Shi1998ParameterOptimization}. The values of the diagonal matrices $\mathbf{R_1}$ and $\mathbf{R_2}$ are drawn from  a uniform distribution $U\sim(0,1)$ each iteration. 

During the PSO iterations, each particle is assessed multiple times on randomized floor patterns employing a fill-ratio of $0.48$, to enhance noise resistance. Initially, all particles are evaluated $N_e=10$ times, resulting in a particle fitness $\mathcal{C}_i$ for particle $i$ defined as the sum of the mean and standard deviation of noise evaluations, as follows:
\begin{equation}
    \label{eq:cost_function_1}
    \mathcal{C}_i = \mu \left( \begin{bmatrix}
        c_1 & c_2 & \hdots & c_{N_e}
    \end{bmatrix}^\top \right) + \sigma \left( \begin{bmatrix}
        c_1 & c_2 & \hdots & c_{N_e}
    \end{bmatrix}^\top \right),
\end{equation}
where $c_j$ represents the fitness obtained from a single simulation. For the top 20\% of the particles, we perform an additional 10 evaluations, bringing the total number of evaluations for these particles to $N_e=20$. Each simulation fitness $c_j$ consists of the sum of errors over robots. Each robot computes its individual error by multiplying two error measures, as outlined in Equation \eqref{eq:cost_function_2}. First, we compute the modeling error $\epsilon_f$ by considering the absolute difference between the estimated fill-ratio $f = \alpha / (\alpha + \beta)$ and true fill-ratio $f^*$, normalized by one tile. Second, we consider a decision error $\epsilon_d$ equal to the normalized decision time of the robot if a correct decision $d_f$ is made. When an incorrect decision is made, we penalize that robot with a factor $\epsilon_d=5$. If by the end of simulation time no decision is made, the normalized maximum decision time is considered ($\epsilon_d=1$). The total error for each robot is the product of $\epsilon_f$ and $\epsilon_d$, and these are summed across all robots to determine the simulation fitness. We use five robots during optimization.
\begin{subequations}
\begin{equation}
    \epsilon_f = 1 + |f - f^*| \cdot (1/N_{\text{tiles}})
\end{equation}
\begin{equation}
    \epsilon_d = \begin{cases}
         t_{d_f}/T_{end} & d_f = d_f^* \\
         5 & d_f \neq d_f^* \\
        1& d_f =-1
    \end{cases}
\end{equation} 
\begin{equation}
     c_j =\sum_{i=1}^{N_{robots}} (\epsilon_f(i) \cdot \epsilon_d(i)) 
\end{equation}
\label{eq:cost_function_2}
\end{subequations}
\section{Experiments and Results} \label{sec:exp_results}
Our results encompass four main elements: (i) calibration, (ii) optimization, (iii) simulated results, and (iv) experimental results.
First, the calibration results include the implementation of stochastic behaviors as outlined in Section \ref{subsec:simulated_environment}, along with the degree of correspondence between the simulated and real environments on key features.
Second, we present the outcomes of the optimization process.
Third, we use the resulting algorithm parameters from optimization to study the performance of our methods in simulation for various fill-ratios, Moran indices, entropy measures, and different numbers of robots. 
Finally, we validate the simulation findings in real experiments. Due to the time-intensive nature of experimental data collection, real experiments are conducted only on a subset of the studies performed in simulation. Specifically, we study all three information sharing strategies in swarm sizes ranging from 5 to 10 robots on a fixed floor pattern with a fill ratio of $0.48$. This constraint underscores the importance of accurate calibration of simulation studies. 
\subsection{Calibration}
\label{sec:calibration}
The modeled stochastic behaviors resulting from (i) mechanical misalignments, (ii) battery drainage, and (iii) observation errors are calibrated using real experimental data collected from five robots over one hour (three sessions of 20 minutes each) in an environment with a fill ratio of 0.48. Each robot in the swarm uses the same parameters for the exploration behavior as $[\gamma_0 \ \gamma \ \tau \ \theta_c] = [7500 \ 15000 \ 2000 \ 50]$. The combined dataset is empirically compared to simulation data from Webots. Iterative adjustments to the simulation parameters are made until proper alignment between simulated and real-world data is achieved. We use cosine similarity to quantify the alignment, as the comparison is conducted on normalized datasets of equal size.
First, the resulting distributions for mechanical misalignment and speed difference are $m_d \sim U(-0.1,0.1)$ and $s_d \sim (\text{Gamma}(3.1,0.095) + 0.8)$, respectively.  Second, modeling the battery voltage drop is done as $b_d \leftarrow \left( 1 - t / (7 \cdot T_{end}) \right)$, set every simulation time step, resulting in initial and final voltage percentage of $100 \%$ and $86\%$ respectively. Third, each vibration sample $\hat E$ is drawn from one of two gamma distributions, specified as:
\begin{equation}
\hat E \sim \begin{cases}
    \text{Gamma}(2.52,0.29) + 0.14 \quad &\text{Vibrating tile} \\
    \text{Gamma}(5.33,0.51) -0.20 \quad &\text{Non-vibrating tile},
\end{cases}
\end{equation}
where we shift the location of the gamma distributions.

\begin{figure}[t]
\centering
\begin{minipage}[b]{1\textwidth}
    \begin{minipage}[b]{0.32\textwidth}
        \centering
        \includegraphics[width=\textwidth]{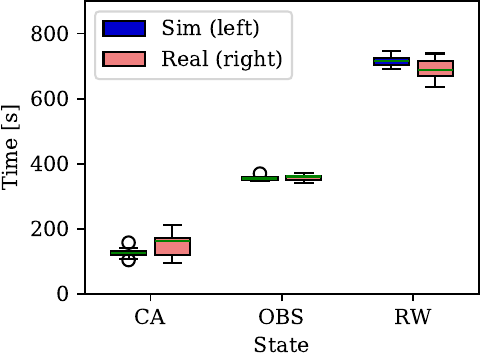}
        (a) State times 
    \end{minipage}
    \begin{minipage}[b]{0.32\textwidth}
        \centering
        \includegraphics[width=\textwidth]{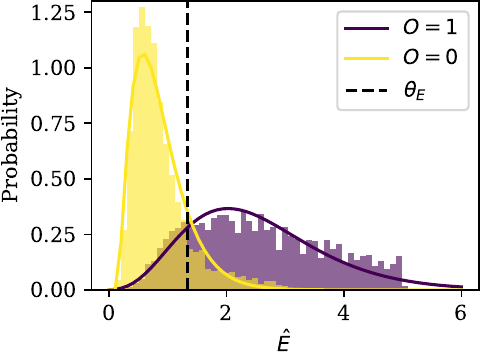}
        (b) Samples, real 
    \end{minipage}
    \begin{minipage}[b]{0.32\textwidth}
        \centering
        \includegraphics[width=\textwidth]{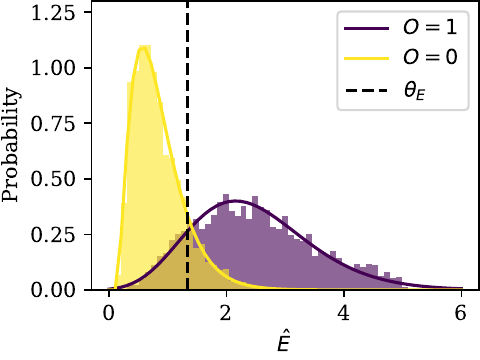}
         (c) Samples, simulation 
    \end{minipage}
\end{minipage}
\caption{Comparison of state times (a) and RMS vibration values $\hat E$ ((b) and (c)) between simulations and experiments. (a) State times obtain a cosine similarity of  $S_i = 0.93$. The observation values $\hat E$ for experimental (b) and simulated (c) data exhibit similar distributions with an equal amount of FP and FN, being around 13\%. The cosine similarities ($S_i$) for $\hat E$ are: 0.97 ($O=1$) and 0.98 ($O=0$).}
\label{fig:calibration_state_and_samples}
\end{figure}

\begin{figure}[t]
\centering 
\begin{minipage}[b]{1\textwidth}
    \begin{minipage}[b]{0.32\textwidth}
        \centering
        \includegraphics[width=\textwidth]{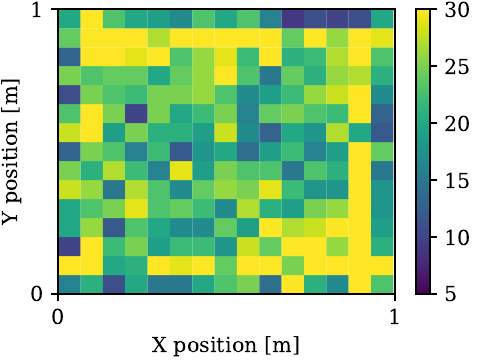}
        (a) Experimental 
    \end{minipage}
    \begin{minipage}[b]{0.32\textwidth}
        \centering
        \includegraphics[width=\textwidth]{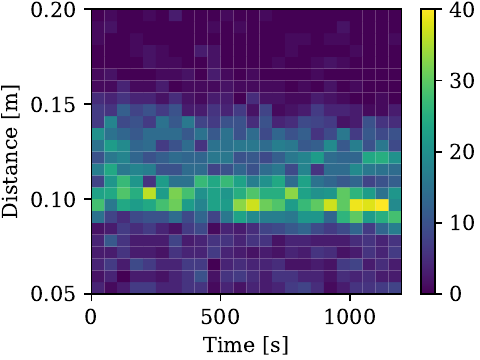}
       (b) Experimental 
    \end{minipage}
    \begin{minipage}[b]{0.32\textwidth}
        \centering
        \includegraphics[width=\textwidth]{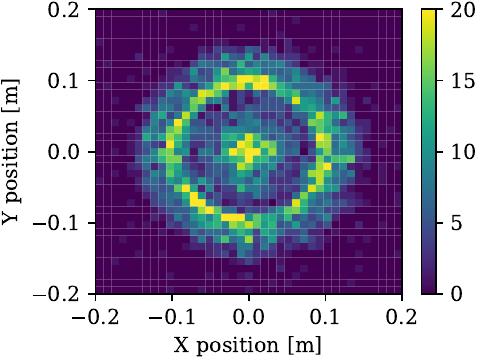}
      (c) Experimental 
    \end{minipage}
\end{minipage}
\centering 
\begin{minipage}[b]{1\textwidth}
    \begin{minipage}[b]{0.32\textwidth}
        \centering
        \includegraphics[width=\textwidth]{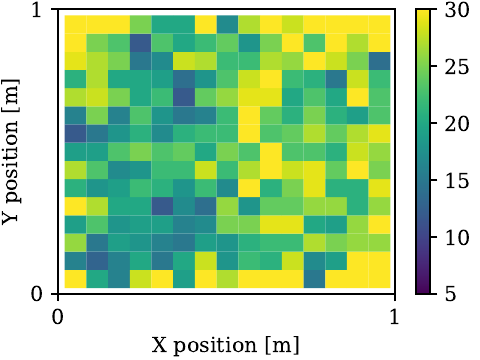}
      (d) Simulation 
    \end{minipage}
    \begin{minipage}[b]{0.32\textwidth}
        \centering
        \includegraphics[width=\textwidth]{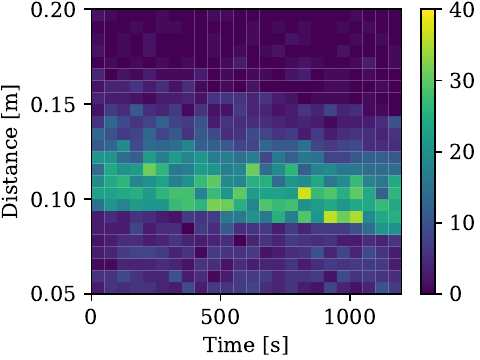}
       (e) Simulation 
    \end{minipage}
    \begin{minipage}[b]{0.32\textwidth}
        \centering
        \includegraphics[width=\textwidth]{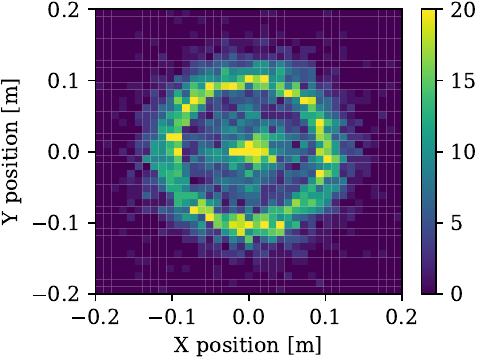}
       (f) Simulation 
    \end{minipage}
\end{minipage}
\caption{Spatial distributions in real and experimental setups. (a) and (d) show the sample distribution, indicating a nearly random spread with higher density in corners due to the incrementation of $\tau$ during collision avoidance. (b) And (e) illustrate the decrease in travel distance over time attributed to battery voltage drop. Finally, (c) and (f) show the distance and angle between consecutive samples, indicating proper randomization of robot-actuation in simulation. The cosine similarities ($S_i$) are: 0.92 ((a) and (d)), 0.92 ((b) and (e)), and 0.89 ((c) and (f))}
\label{fig:calibration_spatial}
\end{figure}

To assess the degree of matching between simulation and real world data, we analyze five key features: (i) time spent in each control state, (ii) observation values, (iii) spatial distribution of samples, (iv) distance traveled between samples over time, and (v) average distance and angle traveled between samples. We use cosine similarity to quantify the similarity $S_i$ between simulation data $X_i$ and real world data $Y_i$ as:
\begin{equation}
    S_i = \frac{X_i \cdot Y_i}{ ||X_i|| \, ||Y_i||}.
\end{equation}
First, the total time spent by the robots is divided into three distinct states: (i) random walk (RW), (ii) sensing (SE), and (iii) collision avoidance (CA). Figure \ref{fig:calibration_state_and_samples}a shows the alignment between simulated state times (left) and experimental state times (right). The overall alignment score obtained equals $S_i = 0.93$, we observe more variance in CA and RW times in the experimental data.
Second, the vibration values $\hat E$ are presented in Figures \ref{fig:calibration_state_and_samples}b and \ref{fig:calibration_state_and_samples}c. The probability of FP and FN are roughly equal in simulation and experiments, measuring 13\%. The obtained similarities are $S_i = 0.97$ and $S_i = 0.98$ for $O=1$ and $O=0$, respectively.
Third, we analyze the spatial distribution of samples, visualized in the 2D histograms in Figures \ref{fig:calibration_spatial}a and \ref{fig:calibration_spatial}d ($S_i = 0.92$). The samples are almost randomly distributed, with higher density near the corners. This effect is caused by the inclusion of CA state time in incrementing $\tau$.
Fourth, the robot's travel distance decreases over time, as shown in Figures \ref{fig:calibration_spatial}b and \ref{fig:calibration_spatial}e ($S_i = 0.92$), following a similar trend in simulation and experiments.
Finally, the distance and angle between two consecutive samples are depicted in Figures \ref{fig:calibration_spatial}c and \ref{fig:calibration_spatial}f ($S_i = 0.89$). 
\subsection{Optimization}
\begin{figure}[t]
    \centering
    \begin{minipage}[b]{0.45\textwidth}
        \centering
        \includegraphics[width=\textwidth]{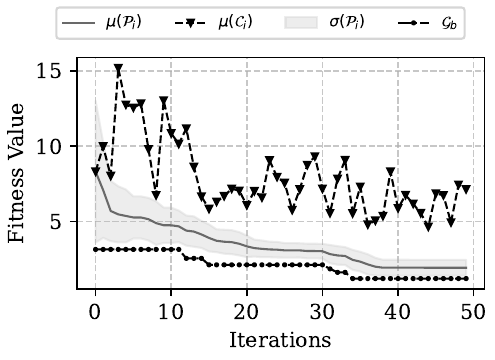}
        (a) Fitness values
    \end{minipage}
    \hspace{0.05\textwidth}
    \begin{minipage}[b]{0.45\textwidth}
        \centering
        \includegraphics[width=\textwidth]{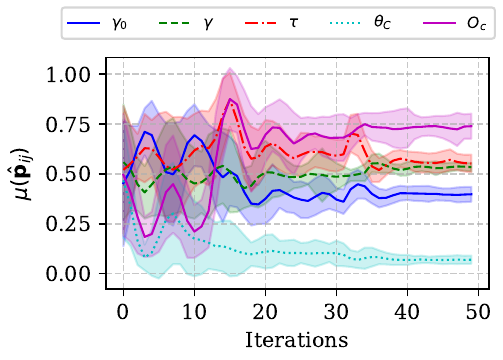}
        (b) Parameter values
    \end{minipage}
\caption{Results of the first stage of optimization: parameter optimization using PSO. (a) Convergence of personal best fitness values \( \mathcal{P}_i \) to the global best fitness \( \mathcal{G}_b \). Average current particle fitnesses are indicated by $\mu(\mathcal{C}_i)$. (b) The average parameter values of all particles over iterations, normalized within the parameter bounds. The parameter values stabilize at around 40 iterations.}
\label{fig:pso_results}
\end{figure}
We consider a two stage approach for optimizing the decision making policy. In the first stage we optimize the algorithm parameters using PSO.
We initialize the parameters randomly within the bounds as specified in Table \ref{table:PSOparameters}. First, the bounds for $\gamma,\gamma_0$ are set empirically to cover a larger space of possible distributions than the size our environment. Second, $\tau$ is limited such that the robot must travel a minimal length of a quarter tile, and a maximum of 1.5 tiles. Third, the minimum value of the collision avoidance threshold is set to 1.5 body lengths. Last, we set an empirically chosen large range for the update count $O_c$. We run 50 iterations with 25 particles.
\begin{table}[b]
\caption{The PSO optimization parameters and bounds. $P_0$ is the empirical best guess particle. $P^*$ is the resulting best particle with respect to our cost-function.}
\label{table:PSOparameters}
\centering
\begin{tabular}{|l|l|l|l|l|l|l|l|}
\hline
\textbf{Parameter} &$\gamma_0$[ms]&$\gamma$[ms]&$\tau$[ms]&$\theta_c$[mm]&$O_c$ \\ \hline
$P_0$         &7500&15000&2000 &50&320 \\ \hline
$\text{min}_i$&0&0&1000 &50&0\\ \hline
$\text{max}_i$&20000&20000&6000&150&500 \\ \hline
$P^*$&7860&10725&3778&55&381  \\ \hline
\end{tabular}
\end{table}
The results of PSO are shown in Figure \ref{fig:pso_results}. In Figure \ref{fig:pso_results}a, the personal best fitness values, denoted as \( \mathcal{P}_i \), converge to the global best fitness value \( \mathcal{G}_b \), indicating convergence. The average fitness values of all particles also decrease over iterations, though with some noise, suggesting that the particles achieve better parameter sets.
Figure \ref{fig:pso_results}b shows the average parameter values across all particles over iterations, with the y-axis normalized within the parameter bounds. Stabilization occurs at around 40 iterations. For the remainder of our simulations and experiments we utilize the global best particle \( P^* = [7860 \ 10725 \ 3778 \ 55 \ 381]^\top \). The final value of $[\gamma \ \gamma_0 ] = [7860 \ 10725]$ suggest a random walk length with high variation and a mean of two tiles. The optimal sampling interval $\tau$ equals 3778ms, corresponding to approximately one tile travel distance, reducing spatial correlation. As expected, the collision avoidance is only triggered when robots are in close proximity, with $\theta_c = 55 \text{mm} \approx 1.5$ body-length. Finally, a minimal number of posterior updates $O_c = 381$ achieves best performance with regard to our fitness function.

\begin{figure}[t]
    \centering
    \begin{minipage}[b]{0.45\textwidth}
        \centering
        \includegraphics[width=\textwidth]{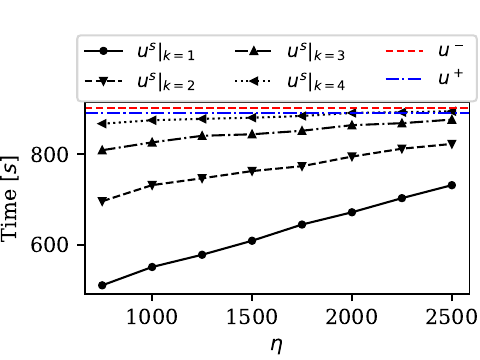}
        (a) Decision time
    \end{minipage}
    \hspace{0.05\textwidth}
    \begin{minipage}[b]{0.45\textwidth}
        \centering
        \includegraphics[width=\textwidth]{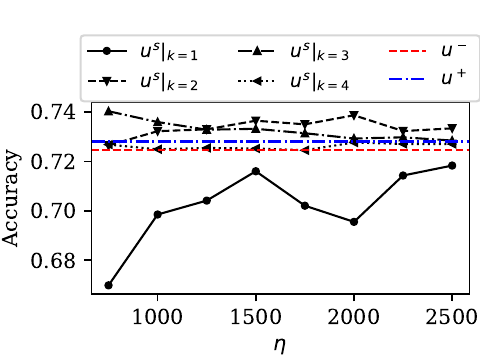}
        (b) Decision accuracy
    \end{minipage}
\caption{Mean performance in 100 randomized simulations for soft-feedback ($u^s$) ($f=0.48$) in terms of time (a) and accuracy (b). The different lines correspond to the values of $\kappa \in \{1,2,3,4\}$. Soft feedback approaches the performance of the originally proposed information sharing strategies $u^-$ and $u^+$ for $\eta \rightarrow \infty, \kappa \rightarrow \infty$.}
\label{fig:calibration_us}
\end{figure}

The second stage of the optimization focuses on tuning the soft-feedback parameters $\eta$ and $\kappa$. We conduct a grid-search that runs a 100 randomized simulations for each combination of $\eta$ and $\kappa$. Empirically, we set the values of interest to $\eta \in \{750,1000,1250,1500,1750,2000,2250,2500 \}$ and $\kappa \in \{1,2,3,4\}$. When $\delta=0$, which happens for $\eta \rightarrow \infty$ and/or $\kappa \rightarrow \infty$, the performance of $u^s$ is equal to that of $u^-$. This is confirmed by our results shown in Figure \ref{fig:calibration_us}. We observe reduced decision times reduce for lower values of $\eta$ and $\kappa$ without losing accuracy. Empirically, we select $\eta = 1500$ and $\kappa=2$ for the remainder of our simulations and experiments, balancing the speed and accuracy of $u^s$.
\subsection{Simulation Results}
\begin{figure}[t]
    \centering
    \begin{minipage}[b]{0.45\textwidth}
        \centering
        \includegraphics[width=\textwidth]{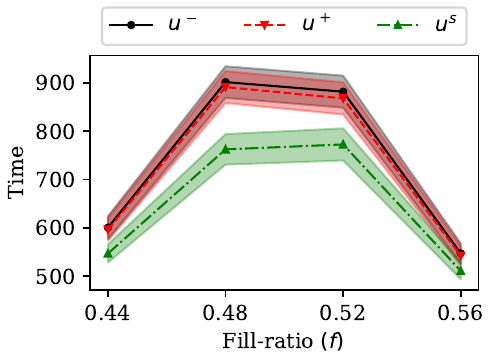}
        (a) Decision time
    \end{minipage}
    \begin{minipage}[b]{0.45\textwidth}
        \centering
        \includegraphics[width=\textwidth]{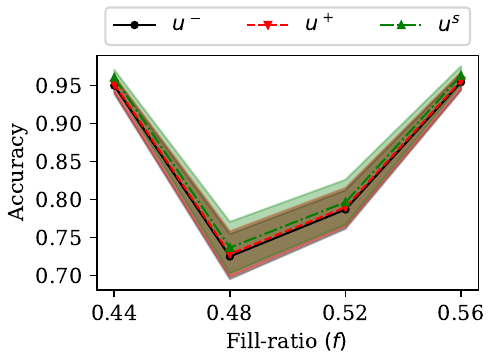}
        (b) Decision accuracy
    \end{minipage}
\caption{Performance of different feedback strategies ($u^-,u^+,u^s$) over 100 randomized simulations with 5 robots for different fill-ratios $f$ in terms of mean and standard error for: time (a) and accuracy (b). We see that soft-feedback decreases decision times without compromising accuracy of decisions.}
\label{fig:fill_ratio_result}
\end{figure}
Our simulation studies consist of three parts: (i) First, we evaluate the performance of $u^-,u^+,u^s$ across different fill-ratios. Second, we assess the scalability of our methods by increasing the number of robots in the environment. Third, we test the robustness of our methods by varying the difficulty of the environment through combinations of fill-ratio, Moran Index, and entropy.

We begin by analyzing the performance of a swarm of 5 robots across the fill ratios $f\in \{0.44,0.48,0.52,0.56\}$. As expected, the problem becomes more challenging at fill ratio near 0.5, as depicted in Figure \ref{fig:fill_ratio_result}. In such hard environments, decision times increase, and decision accuracies decrease. Regarding the different feedback strategies, soft feedback reduces decision times without compromising accuracy. This decrease in decision time is particularly notable at the more difficult fill ratios. This demonstrates the effectiveness of soft-feedback in challenging environments, with a reduction in decision times of approximately \(17\% \  ((900-750)/900\cdot 100\%)\). All information sharing strategies exhibit a similar spread in decision time and decision accuracy.
\begin{figure}[t]
    \centering
    \begin{minipage}[b]{0.45\textwidth}
        \centering
        \includegraphics[width=\textwidth]{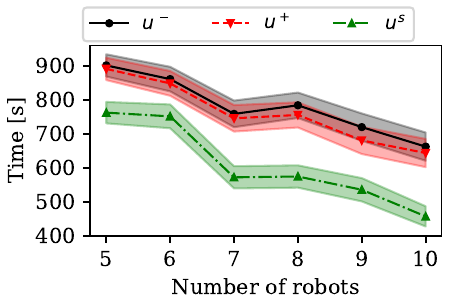}
        (a) Decision time
    \end{minipage}
    \hspace{0.01\textwidth}
    \begin{minipage}[b]{0.45\textwidth}
        \centering
        \includegraphics[width=\textwidth]{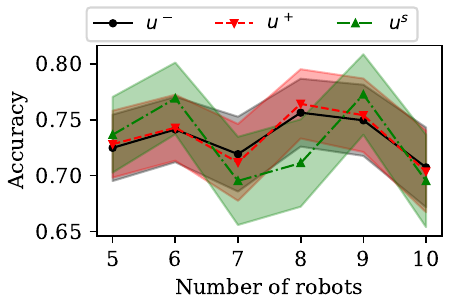}
        (b) Decision accuracy
    \end{minipage}
    \begin{minipage}[b]{0.45\textwidth}
        \centering
        \includegraphics[width=\textwidth]{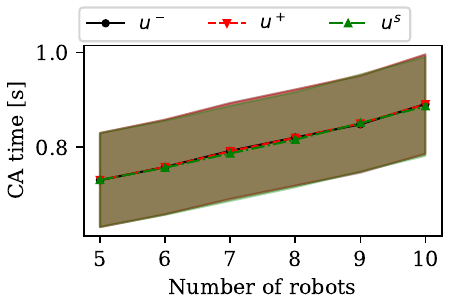}
        (c) CA time per sample
    \end{minipage}
    \hspace{0.01\textwidth}
    \begin{minipage}[b]{0.45\textwidth}
        \centering
        \includegraphics[width=\textwidth]{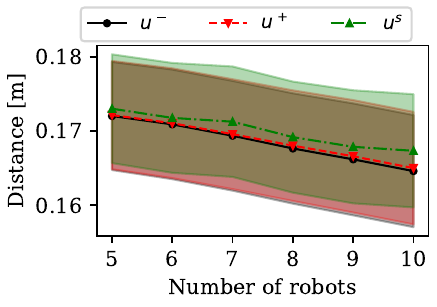}
        (d) Intersample distance
    \end{minipage}
\caption{Mean performance of different information sharing strategies ($u^-,u^+,u^s$) over 100 randomized simulations ($f=0.48$) for different swarm sizes in terms of: decision time (a), decision accuracy (b), collision avoidance time per sample (c), and distance traveled between samples (d). As the swarm size increases, decision times decrease. However, accuracy does not increase proportionally. As a result of the increasing CA time per sample, spatial correlation of samples in increased.}
\label{fig:multi_robot_result}
\end{figure}
\begin{table}[b]
\caption{The difficulty of the environments used in simulations in which we compare optimized and empirical algorithm parameters. Difficulty is expressed in terms of Moran Index ($E_{MI}$) and Entropy ($E_e$) values for fill-ratios $0.48$ and $0.46$}
\label{table:MI_Entropy}
\centering
\begin{tabular}{|l|l|l|l|l|l|l|l|}
\hline
\textbf{Pattern} &Diagonal&Stripe&Block Diagonal&Alternating&Random \\ \hline
$E_{MI}|_{f=0.48}$  &0.73&0.88&0.76&-0.96&0.00 \\ \hline
$E_e|_{f=0.48}$      &1.0&1.0&0.85&0.0&0.51\\ \hline
$E_{MI}|_{f=0.46}$    &0.71&0.88&0.75&-0.91&-0.05 \\ \hline
$E_e|_{f=0.46}$       &0.99&0.99&0.84&0.0&0.46\\ \hline
\end{tabular}
\end{table}
\begin{figure}[t]
    \centering
    \includegraphics[width=0.99\linewidth]{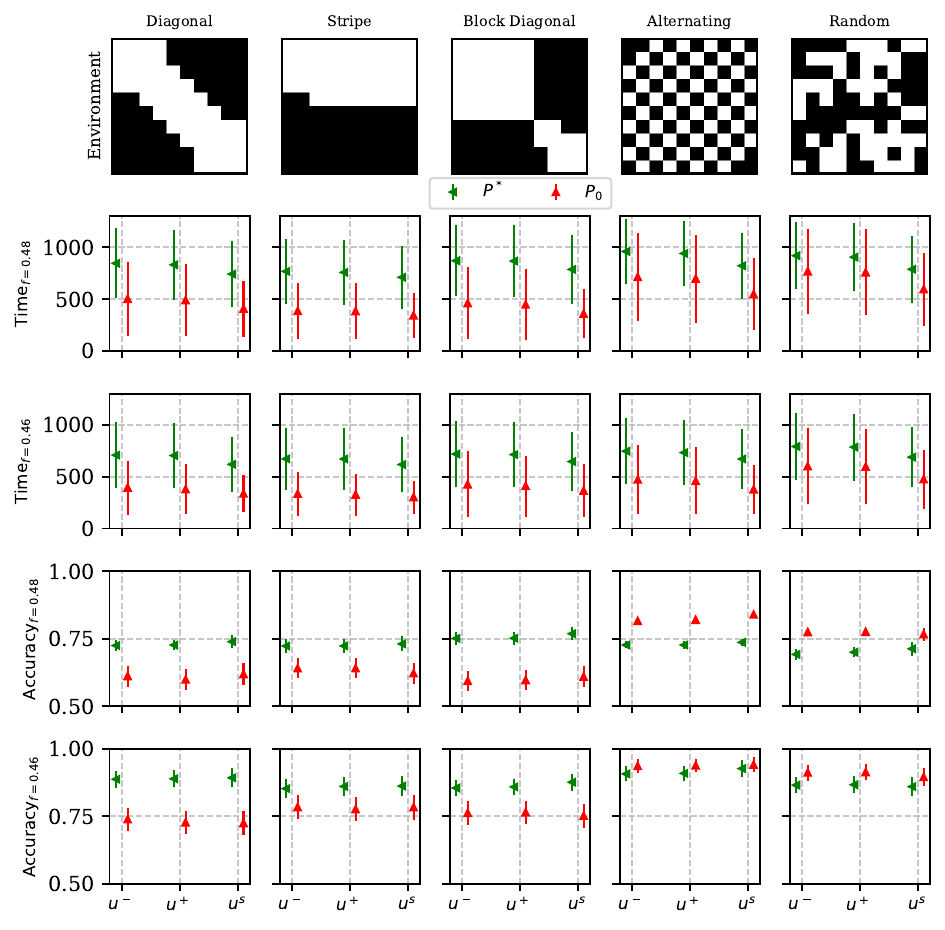}
    \caption{Decision time and decision accuracy results are presented across four rows. Row 1 and 2 display the mean and standard deviation of decision time, row 3 and 4 show the mean and standard error of decision accuracy for enhanced readability. We compare the optimized parameter set \( P^* \) against the empirical parameter set \( P_0 \) at fill ratios $f=0.48$ and $f=0.46$. The results show that the optimized parameter set \( P^* \) has consistent performance across patterns of varying difficulty. In contrast, the empirical parameter set \( P_0 \) performs well on easier patterns (column 4-5) but exhibits a significant decline in accuracy on more challenging patterns (column 1-3).}
    \label{fig:environment_changes}
\end{figure}

Second, we examine the scalability of our methods by increasing the swarm size. The decentralized control approach enables easy scaling of the swarm size, resulting in an increased information rate. Hypothetically, this leads to reduced decision times, and increased accuracy, due to the spread over robots over the environment. However, robots can also become cluttered, increasing spatial correlation, thus reducing sampling quality. The latter is confirmed by our results in Figure \ref{fig:multi_robot_result}, where we show the decision time, decision accuracy, time spent in collision avoidance, and average distance driven between consecutive samples. As expected, decision times decrease with larger swarm sizes, but accuracy does not improve. In fact, accuracy may even decline due to increased time spent in collision avoidance (Figure \ref{fig:multi_robot_result}c). During collision avoidance, robots turn random angles, reducing the distance traveled between consecutive samples (Figure \ref{fig:multi_robot_result}d). This shorter travel distance increases spatial correlation in the collected samples, as robots are more likely to revisit areas, reducing the representativeness of local samples in modeling the overall fill ratio.

Third, we assess the performance of our methods and optimization by placing a swarm of five robots in environments of varying difficulty. The environment is reduced to a $10\times10$cm grid, resulting in $100$ tiles. This decrease in tile size allows for accurate construction of specific floor patterns. We evaluate decision time and decision accuracy over 100 randomized simulations in the environments: diagonal, stripe, block diagonal, alternating, and random at fill ratios of $0.48$ and $0.46$. The corresponding difficulties in terms of Moran Index and Entropy are listed in Table \ref{table:MI_Entropy}. We compare our optimized algorithm parameters ($P^*$) against the empirical parameters ($P_0$), which served as the best initial guess in the PSO process. The results, shown in Figure \ref{fig:environment_changes}, show that the optimized particle performs consistently across all environments. In contrast, the empirical particle performs well in easier environments (random and alternating) but experiences a significant accuracy drop in more challenging environments (diagonal, stripe, and block diagonal). These results highlight that optimizing algorithm parameters across a variety of environments (noise-resistant optimization) can provide consistent accuracy of decisions in various environments.

\subsection{Experimental Results}\label{sec:real_robot_results}
\begin{figure}[t]
    \centering
    \begin{minipage}[b]{1\textwidth}
        \centering
        \includegraphics[width=0.9\textwidth]{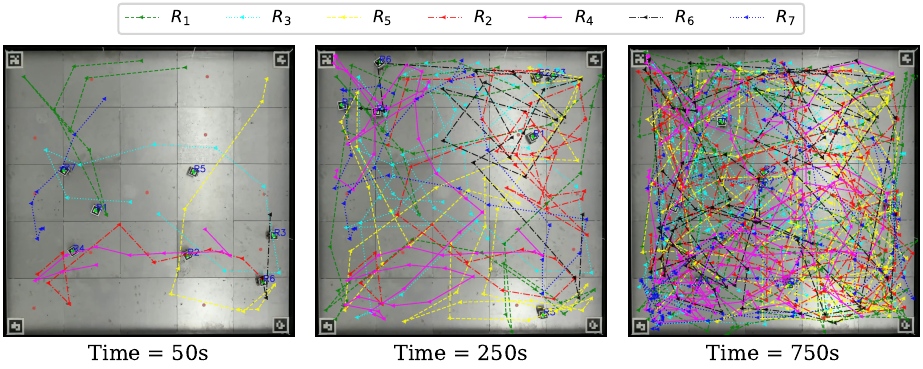}
    \end{minipage}
    \caption{An example of the paths traveled and the samples taken in an experiment with 7 robots. The robots randomly acquire samples to accurately estimate the surface's fill ratio.}
    \label{fig:exp_example}
\end{figure}

We validate our simulation findings in real experiments using swarm sizes of: $\text{size} \in \{5,6,7,8,9,10\}$. We use robots from a population of 16 available robots. To ensure a fair comparison, the strategies \( u^-, u^+, \) and \( u^s \) are implemented concurrently, allowing all information-sharing strategies to utilize the same observations. Consequently, communication is implemented using three binary messages, one corresponding to each communication strategy. All experiments are run for a maximum of 1000 seconds, or until all robots reach a decision for all strategies. We use an environment with a fill-ratio $f=0.48$. An example of the robots' coverage is shown in Figure \ref{fig:exp_example}. We evaluate our experiments on (i) decision time, (ii) decision accuracy, (iii) collision avoidance times, and (iv) package loss within the network in real world conditions.
\begin{figure}[t]
    \centering
    \begin{minipage}[b]{1\textwidth}
        \centering
        \includegraphics[width=0.95\textwidth]{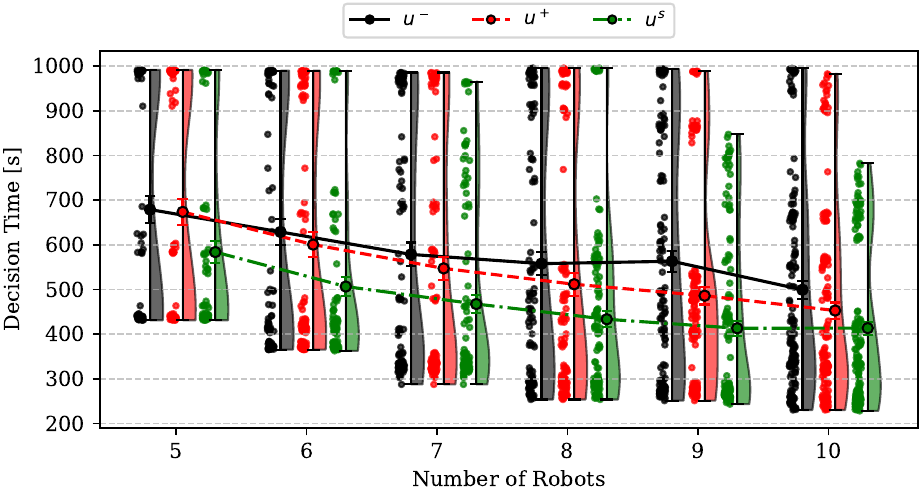}
    \end{minipage}
    \caption{Decision times for each robot across 15 experiments for swarm sizes in $\{5, 6, 7, 8, 9, 10\}$. Decision times are scattered alongside their probability distributions. The line and error bars indicate means and standard errors for different feedback strategies. The results indicate that the soft feedback strategy ($u^s$) consistently outperforms both $u^-$ and $u^+$ in decision time. Unlike in simulations, $u^+$ outperforms $u^-$ as the swarm size increases. This behavior is almost certainly caused by increased packet loss in the network as swarm sizes increase, allowing one robot to push the swarm to consensus using of positive feedback.}
    \label{fig:exp_decision_times}
\end{figure}

\begin{figure}[t]
    \centering
    \begin{minipage}[b]{0.49\textwidth}
        \centering
        \includegraphics[width=\textwidth]{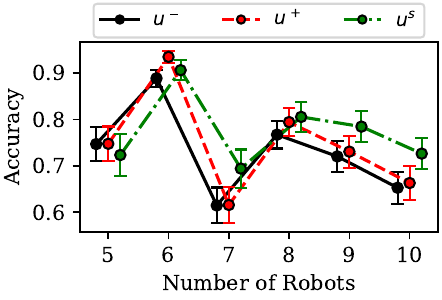}
        (a) Decision Accuracy
    \end{minipage}
    \begin{minipage}[b]{0.49\textwidth}
        \centering
        \includegraphics[width=\textwidth]{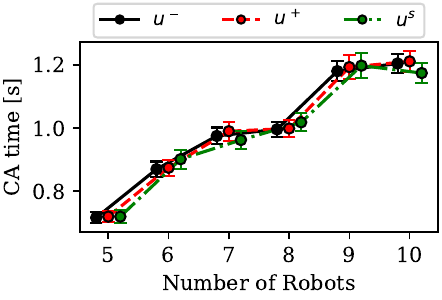}
        (b) CA time per sample
    \end{minipage}
    \begin{minipage}[b]{0.49\textwidth}
        \centering
        \includegraphics[width=\textwidth]{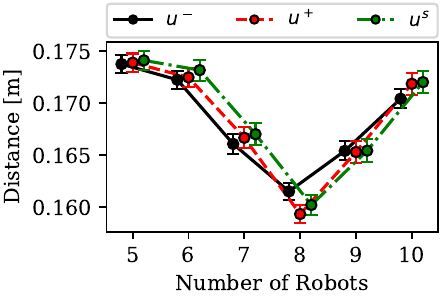}
        (c) Intersample distance
    \end{minipage}
    \begin{minipage}[b]{0.49\textwidth}
        \centering
        \includegraphics[width=\textwidth]{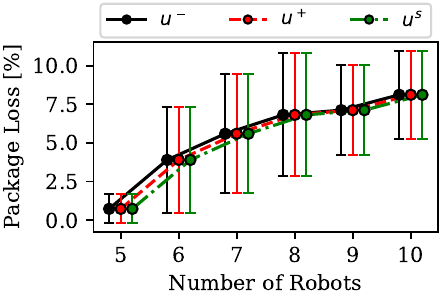}
        (d) Package loss
    \end{minipage}
\caption{Mean values for different feedback strategies ($u^-,u^+,u^s$) over 15 experiments ($f=0.48$) for different swarm sizes in terms of decision accuracy (a), collision avoidance time per sample (b), distance between samples (c), and network loss (d). As the swarm size increases, accuracy does not grow proportional, and packet loss is increased. As a result of the increasing CA time per sample, exploration becomes less accurate.}
\label{fig:exp_results}
\end{figure}

The decision time of individual robots over various swarm sizes is shown in Figure \ref{fig:exp_decision_times}. Similar as in simulations, $u^s$ outperforms the strategies $u^-$ and $u^+$. The difference in observed decision times in reality and simulation can be attributed to the fact that, in the experiments, we use a single floor pattern with multiple initial robot placements, whereas in the simulations, both the floor pattern and initial placements are randomized. The faster decision times of \( u^+ \) compared to \( u^- \) can be attributed to network packet loss, which increases approximately linearly with swarm size, ranging from \( 0\% \) to \( 7.5\% \), as shown in Figure~\ref{fig:exp_results}d. The observed loss breaks all-to-all communication and causes discrepancies in robots' beliefs. This allows positive feedback to push the swarm to consensus when a single robot reaches a decision. Despite the fact that all-to-all communication is not available anymore, our soft feedback approach showed faster decision times, showing its robust effect in uncertain communication structures. 
Reason for the observed network packet loss is that, for RF24 communication, we use the the well-established open-source repository by \citet{TMRh202023RF24Network}. This implementation follows an OSI Network Layer structure, creating hierarchies among network nodes. When communication is structured to be flat, this hierarchy can lead to certain nodes becoming more 'busy' as they relay messages to their children, and vice versa. Additionally, the minimalistic hardware of the robots limits their ability to continuously monitor the network while performing tasks like sensing.
When looking into the accuracy of decisions, shown in Figure \ref{fig:exp_results}a, similar observations are made as in simulations. The accuracy does not grow proportional to swarm sizes. As expected, the collision avoidance times increase as the swarm size grows, as shown in Figure \ref{fig:exp_results}b. Examining the resulting distance between samples, we observe that it decreases until the swarm size exceeds 8 robots (Figure \ref{fig:exp_results}c). The increase in distance for swarm sizes larger than 8 robots is due to using only a subset of the 16 available robots, while experiments with 5–8 robots used all 16 in multiple batches. This reduces representativeness for experiments with more than 8 robots, as parameters like robot speed may vary within the selected subset.

\section{Discussion and Conclusion}
Robotic swarms hold great potential for distributed sensing and inspection applications. In this study, we investigated the use of mobile robots as sensor carriers for automated vibration-based inspections. 
To support this vision, we examined a binary decision making problem in a tiled environment of 25 tiles that are either vibrating or non-vibrating. The goal of the swarm is to determine the majority tile type. In our approach, we used a Bayesian decision-making algorithm from existing literature. This algorithm employs a Beta distribution to model the proportion of vibrating tiles in the overall setup. The Beta distribution is updated following a Bayesian approach and based on (i) a robot's individual observations and (ii) the information of other robots shared within the swarm. In particular, we studied three information sharing strategies: (i) no feedback ($u^-$), where robots continuously broadcast their individual observations \citep{Ebert2020BayesSwarms}. (ii) Positive feedback ($u^+$), where robots initially broadcast their individual observations before they reach a final decision, and switch to broadcasting their final decisions upon reaching one \citep{Ebert2020BayesSwarms}. (iii) Our newly proposed strategy soft feedback ($u^s$), in which robots broadcast randomized samples proportional to the robots' belief and current observation \citep{Siemensma2024CollectiveInspection}. We assessed the performance of all three strategies in simulation and real-world experiments, using real vibration sensing robots. Our simulation framework employs the physics-based robotic simulator Webots, and is calibrated to our real experimental setup.
The decision making problem difficulty lies in the environment and the robots behavior. We optimized the algorithm parameters that govern robots behavior by leveraging our simulation framework. We used a two-stage optimization approach: first we used a Particle Swarm Optimization (PSO) method to optimize common algorithm parameters across the three feedback strategies, we then used a grid search to optimize the soft feedback parameters. 
We evaluated all three information sharing strategies in simulation. Our simulation results indicate that soft feedback outperforms no feedback and positive feedback in terms of decision times without compromising accuracy. This result is consistent over the different swarm sizes and environmental difficulties which we studied in this work. Moreover, our optimized parameter set significantly outperforms empirically chosen parameters in difficult environments, showing the necessity of optimization for consistent and accurate decision making.
Our experimental findings are in line with simulated findings, showing the robustness of soft feedback in real world conditions. The results presented in this work demonstrate that the difficulty in collective perception arises from both robot behaviors and environmental difficulties. This finding aligns with the literature \citep{Ebert2020BayesSwarms,Valentini2016CollectiveSwarm,Bartashevich2019BenchmarkingDecision-making}. Additionally, we present a novel optimization framework calibrated to real robot behaviors, a crucial step before deploying swarms in real-world environments outside of simulation \citep{Francesca2016AutomaticChallenges,Brambilla2013SwarmPerspective}. Overall, this work highlights a first step towards the deployment of collective decision making policies in real world sensing and inspection scenarios using miniaturized robot swarms.

 The experimental validation of the performance of various information sharing strategies show similarities with our simulated findings. However, positive feedback progressively performed better than no feedback with increasing swarm sizes, a phenomenon absent in simulation. This increasing difference is likely a result of increased network packet loss as the gap appears to scale proportionally with network loss. This finding is consistent with \citet{Ebert2020BayesSwarms}, which studied non-all-to-all communication scenarios within the swarm.
Additionally, our proposed soft feedback approach returned superior decision times for all swarm sizes. However, to obtain such effective behavior, soft feedback requires optimization. Poor parameter tuning can lead to significantly faster but inaccurate decisions (e.g., $\kappa=1$) or performance similar to the no feedback strategy (e.g., $\kappa=4$). The optimal parameter set for soft feedback depends on the swarm size, environmental difficulty, and the desired decision time and accuracy of the designer. In addition, the fact that our optimized parameters significantly outperform empirical ones in complex environments, underscores the need for optimization of decision making strategies. 
Our optimization framework offers a tool for such optimization within physics-based simulations. While the optimization yields successful robot behaviors, it does not account for all possible parameters (e.g., swarm size). Presumably, the performance of decision making improves when each swarm size is optimized individually or when swarm size is explicitly accounted for during the optimization process. However, using our current optimization framework, this would lead to large computation times due to the non-linear increase in required particles and iterations when increasing dimensions in PSO. Similarly, our calibration represents a valuable step towards modeling real world robot behaviors. However, the approach we took in this work has been largely empirical, with features and parameters selected based on observations in experiments, which may hinder its generalizability. For instance, our calibrated simulation did not account for network loss, a limitation that became apparent during experiments with larger swarm sizes ($\text{size} > 5$) compared to the size used during calibration ($\text{size} = 5$). These findings highlight the need for a more generally calibrated framework to reliably optimize robot behaviors when designing collective decision making policies in real world scenarios. 

While this study shows the potential of vibration based inspections using miniaturized robot swarms, its direct application to inspection of infrastructure is still limited. For real-world inspection and monitoring, both hardware and software require further development. Currently, our robot's IMU has a limited sensing capability with a sampling rate of 350Hz, while sensors used in industry can reach up to 6000Hz \citep{Amazon2024AmazonMonitron}. Additionally, our onboard processing uses a simplified vibration-magnitude extraction, whereas real world applications require extracting more complex signal features, such as eigenfrequencies, to detect damage \citep{Farrar2007AnMonitoring}. To extract such complex signal features, the robot's dynamics must additionally be considered. Similarly, the implementation of our communication using the RF24 library imposes limitations on studying various network structures. To further advance our research into different network structures, a more advanced implementation must be developed, which can then be accurately accounted for within the calibration process.

In future work, we aim to extend the contributions of this paper in several ways. First, we plan to enhance our calibration and optimization framework. Specifically, pre-identifying algorithm parameters most influential to the task outcome significantly reduces the search space in optimization. Leveraging optimization methods like Bayesian Optimization (BO) can minimize the number of function evaluations compared to PSO.
This is especially relevant for computationally expensive simulations.
Second, new directions of research for miniaturized vibration sensing robots include deploying them in increasingly challenging environments and more complex inspection tasks. These tasks can include environments containing multiple tile types with time dependent properties. Robots can model and detect these time-dependent properties, which would prompt them to reset or adjust their current belief. Furthermore, enhancing the hardware to detect signal features relevant to infrastructural monitoring offers a crucial step toward the successful deployment of robot swarms in real-world scenarios. Finally, we aim to enhance our RF24 communication system to better support swarm-like communication.

\bmhead{Acknowledgements}

The authors wish to thank Simon Busman, Taraneh Mokabber, and Sepide Taleb for their contributions to creating the experimental setup, Hans J.G. Boersma for his work on creating high-resolution images from our robot, and Darren Chiu, Sneha Ramshanker, and Radhika Nagpal for their contributions to the work presented in the earlier conference manuscript that lays the foundation of this work.  

 \bibliography{references}









\end{document}